\documentclass[10pt,journal,compsoc]{IEEEtran}

\usepackage{cite}
\usepackage{amsmath,amssymb,amsfonts}
\usepackage{graphicx}
\usepackage{caption}
\usepackage{textcomp}
\usepackage{xcolor}
\def\BibTeX{{\rm B\kern-.05em{\sc i\kern-.025em b}\kern-.08em
    T\kern-.1667em\lower.7ex\hbox{E}\kern-.125emX}}
\usepackage{amsmath}
\usepackage{algorithm}
\usepackage[noend]{algpseudocode}
\usepackage{float} 
\usepackage{subfigure}
\usepackage{multirow}
\usepackage{mathrsfs}
\usepackage{CJK}
\usepackage{hyperref}

\newcommand{\algo}{\textsf{FinGAT}}
\newcommand{\algorelation}{\textsf{FinGAT-NT}}

\newcommand{\ct}[1]{\textcolor{red}{#1}}
\newcommand{\comment}[1]{}

\begin{document}
%
\title{FinGAT: Financial Graph Attention Networks for Recommending Top-$K$ Profitable Stocks
}
%
%
%
%

\author{Yi-Ling Hsu\textsuperscript{\IEEEauthorrefmark{1}}, Yu-Che Tsai\textsuperscript{\IEEEauthorrefmark{1}},
        Cheng-Te Li,~\IEEEmembership{Member,~IEEE}
\IEEEcompsocitemizethanks{
\IEEEcompsocthanksitem Yi-Ling Hsu, Master Program in Statistics, National Taiwan University, Taipei, Taiwan.
\IEEEcompsocthanksitem Yu-Che Tsai, Department of Computer Science and Information Engineering, National Taiwan University, Taipei, Taiwan.
\IEEEcompsocthanksitem Cheng-Te Li, Institute of Data Science, National Cheng Kung University, Tainan, Taiwan.
}
\thanks{Manuscript received OOOO OO, 20OO; revised OOOO OO, 20OO.}}

%
%

\markboth{IEEE Transactions on Knowledge and Data Engineering (TKDE) 2021}%
{Shell \MakeLowercase{\textit{et al.}}: Bare Advanced Demo of IEEEtran.cls for IEEE Computer Society Journals}
%



\IEEEtitleabstractindextext{%
\begin{abstract}
Financial technology (FinTech) has drawn much attention among investors and companies. While conventional stock analysis in FinTech targets at predicting stock prices, less effort is made for profitable stock recommendation. Besides, in existing approaches on modeling time series of stock prices, the relationships among stocks and sectors (i.e., categories of stocks) are either neglected or pre-defined. Ignoring stock relationships will miss the information shared between stocks while using pre-defined relationships cannot depict the latent interactions or influence of stock prices between stocks.
In this work, we aim at recommending the top-K profitable stocks in terms of return ratio using time series of stock prices and sector information. We propose a novel deep learning-based model, Financial Graph Attention Networks (FinGAT), to tackle the task under the setting that no pre-defined relationships between stocks are given. 
The idea of FinGAT is three-fold. First, we devise a hierarchical learning component to learn short-term and long-term sequential patterns from stock time series. Second, a fully-connected graph between stocks and a fully-connected graph between sectors are constructed, along with graph attention networks, to learn the latent interactions among stocks and sectors. Third, a multi-task objective is devised to jointly recommend the profitable stocks and predict the stock movement. Experiments conducted on Taiwan Stock, S\&P 500, and NASDAQ datasets exhibit remarkable recommendation performance of our FinGAT, comparing to state-of-the-art methods.
\end{abstract}

}

\maketitle

\maketitle
\begingroup\renewcommand\thefootnote{\IEEEauthorrefmark{1}}
\footnotetext{Equal contribution}
\endgroup

\IEEEdisplaynontitleabstractindextext

%
\IEEEpeerreviewmaketitle

\ifCLASSOPTIONcompsoc
\section{Introduction}
\label{sec-intro}


The stock market has grown swiftly in these years, and trading stocks have become one of the most attractive financial instruments for investors. Investing in the stock market is highly profitable and easy to get started. 
However, investing stocks usually involves extremely high risk, which makes drawing up a proper investment plan a crucial task. Previously, people tend to empirically choose stocks by their financial knowledge or expertise. As financial technology (FinTech) is now in widespread use, people come up with statistical inference models to forecast the dynamic movement of stock prices~\cite{sezer2020financial}. Techniques of machine learning and deep learning are investigated and applied in industries, which has shown remarkable success in different stock markets, such as S\&P 500~\cite{kim2019hats} and NASDAQ~\cite{feng2018enhancing}. 
\begin{figure}[!t]
  \centering

  \includegraphics[width=1.0\linewidth]{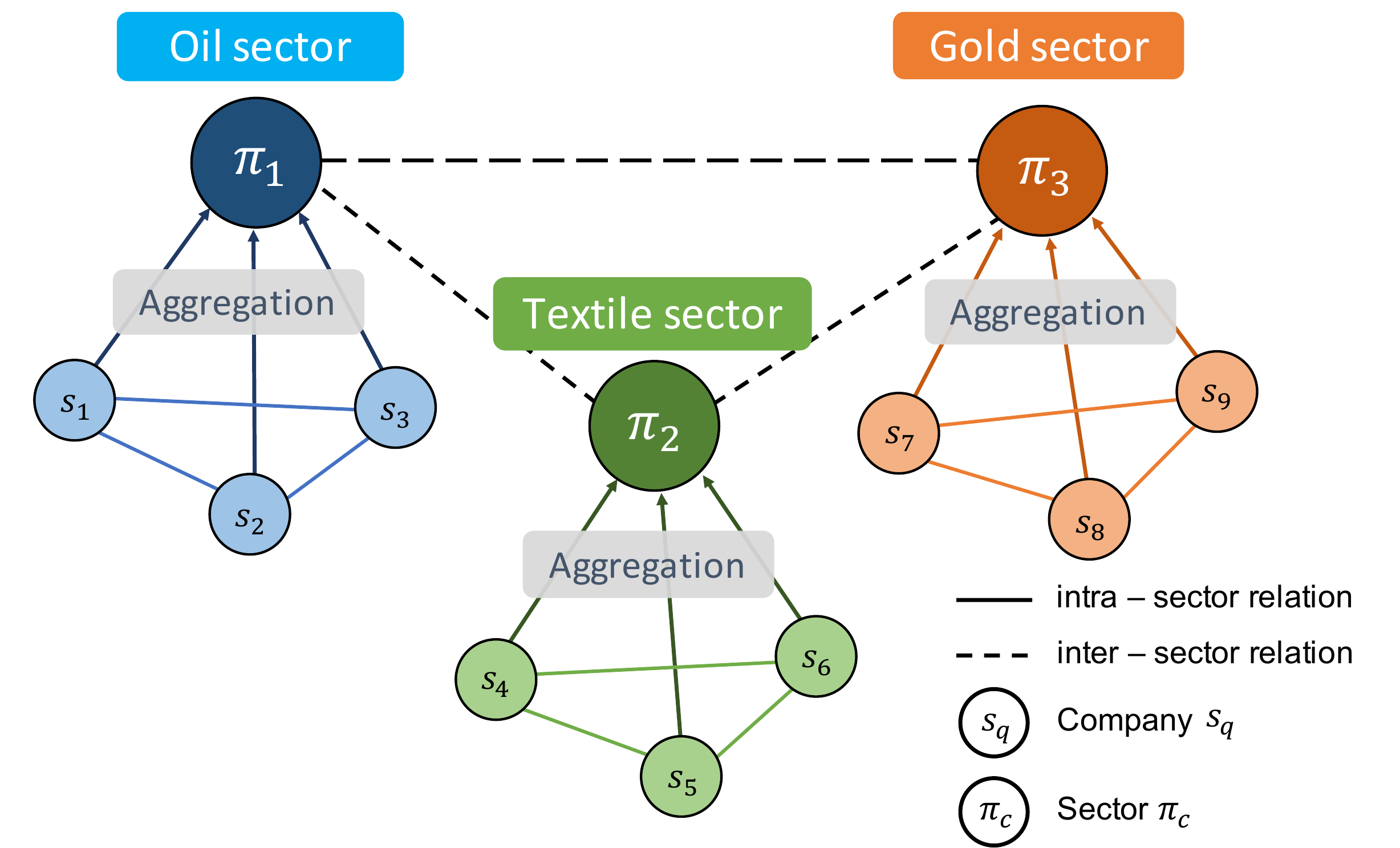}
  \caption{A toy example of intra-sector relations and inter-sector relations.}
  \label{fig:Toy_example}
\end{figure}

In predicting stock prices, typical methods such as Auto Regression-based methods~\cite{box2015time,engle1982autoregressive} treat time series indicators (e.g., stock price) as a linear stochastic process.
However, a stock time series usually appears in a dynamic nonlinear process. Regarding this drawback, deep learning methods such as Recurrent Neural Network (RNN)~\cite{baek2018modaugnet,chen2015lstm} project time series into a high-dimensional space to obtain its sequential-level representation. Attention mechanism~\cite{li2018stock} has been applied to consider the varying importance of each timestamp by giving learnable attentive weights. 
With the rise of graph neural network, recent studies~\cite{chen2018incorporating,kim2019hats} incorporate the relationships (e.g., upstream, downstream, and sector) between stocks to form graphs, which are further used to pass financial knowledge between stocks so as to distill graph-level features.
A key idea is that the stock price is not only affected by the company itself, but also determined by the global trend of the market or the financial situation of its competitors.
Hence, modeling with graph neural networks seamlessly unifies the information of the targeted company as well as the correlated companies based on the constructed graphs.
In contrast with stock price prediction, since the investors care more about identifying stocks that can bring higher return in the future, few studies have attempted to recommend profitable stocks.
Convolutional and recurrent neural networks are effective in extracting long-term and short-term sequential features from the time series of stock prices and producing promising recommendation performance~\cite{feng2019temporal,tsai2019finenet}.
Nevertheless, it is still worthwhile to study how to model the relationships between stocks for profitable stock recommendation. 

It is challenging to model the relationships between stocks (i.e., listed companies)\footnote{Terms ``stocks'' and ``listed companies'' are used interchangeably throughout this paper.} in recommending profitable stocks. First, the data on company-company relationships, such as ``investing'', ``member of'', ``subsidiary'', and ``complies with'', is difficult to access due to confidentiality agreement, security issues, or privacy concerns. Manually collecting the relationship data could be either incomplete or labeling-bias. Second, the relationships between listed companies are dynamic. Two companies can change their relationships between competition and support with time. It is less possible to track the evolution of all their relationships. Third, since the latent relationships between \textit{sectors} (i.e., stock categories), such as the underlying correlation among ``oil'', ``textile'', and ``gold'' sectors, can also affect the rise and fall of stock prices. Such \textit{inter-sector relations} are usually implicit and hard to be concretely defined, comparing to the explicit company-company relations, which are termed \textit{intra-sector relation}. 

We use Figure~\ref{fig:Toy_example} as a toy example to illustrate the intra-sector relations between stocks and inter-sector relations between sectors, which can be regarded as \textit{hierarchical} relationships. Same-color company nodes $s_q$ are stocks belonging to the same sector $\pi_c$. Solid lines indicate intra-sector relations while dashed lines refer to inter-sector relations. 
In the real world, stocks within a sector (e.g., oil) usually have a similar price movement trend. Prices of related sectors (e.g., gold and textile) can be influenced by stock prices in the oil sector. For example, the growth of oil prices usually involves the happening of inflation phenomena and leads to the increasing uncertainty of economic development.
Since gold is a financial item against inflation, the demand for gold tends to increase, which is followed by the growth of oil prices. Such an example exhibits the dependency of inter-sector relationship between oil and gold sectors~\cite{zhang2010crude}. 
Nevertheless, it is difficult to represent such prior knowledge on which sector-sector pairs are with high dependency in stock price movement.

In this paper, we aim at recommending the most profitable stocks. Given the historical time-series data of stock prices for a set of listed companies, our goal is to recommend stocks that can bring the highest return by investing them in the next day. To better represent each stock, we will model not only the sequential patterns hidden in time series, but also the hierarchical influence between stocks at both stock and sector levels. That said, we aim to learn how stocks are influenced by and interacted with each other through modeling the latent relationships between stocks, between sectors, and between stocks and sectors, as depicted in Figure~\ref{fig:Toy_example}, based on features extracted from times series of historical stock prices.

We propose a novel graph neural network-based model, \textbf{\underline{Fin}}ancial \textbf{\underline{G}}raph \textbf{\underline{A}}ttention Networks (\algo), to achieve the goal and to implement our idea by dealing with the aforementioned challenges in modeling hierarchical relationships among stocks and sectors. The proposed \algo~consists of three main phases, \textit{stock-level feature learning}, \textit{sector-level feature learning}, and \textit{multi-task learning}. First, we extract a variety of features to represent every stock at a day, and exploit attentive gated recurrent units (GRU) to learn short-term (i.e., single-week) sequential features. By constructing a fully-connected graph of stocks belonging to the same sector, we utilize Graph Attention Network (GAT)~\cite{velivckovic2017graph} to learn their latent intra-sector relations. Second, we learn long-term sequential features by creating a weekly aggregation layer that combines each stock's short-term embeddings through attentive GRU. In addition, a graph pooling mechanism is proposed to generate the long-term embeddings of same-sector stocks to the corresponding sector embedding. GAT is again applied to learn the latent inter-sector relations. Third, since the profit of a stock is influenced by two highly-correlated factors, stock price return (real value) and stock movement (binary value), we devise a multi-task learning method to jointly optimize such two tasks. The predicted return values are used to generate the ranking of stocks for top-$K$ profitable stock recommendations.

We summarize the contributions of our work as follows.
\begin{itemize}
    \item Conceptually, we propose to recommend the most profitable stocks by modeling the hierarchical correlation of stocks, including intra-sector relations, inter-sector relations, and stock-sector relations. Comparing to existing studies that use pre-defined knowledge of stock-stock relations, it is novel to learn such latent relations for stock recommendation in this work. 
    \item Technically, we devise a novel multi-task graph neural network-based model, \algo~\footnote{The code of \algo~can be accessed via the following Github link: \url{https://github.com/Roytsai27/Financial-GraphAttention}}, to fulfill our idea and generate the most profitable stocks. \algo~is able to learn sequential patterns in financial time series and hierarchical influence among stocks and sectors.
    \item Empirically, experiments conducted on Taiwan Stock, S\&P 500, and NASDAQ datasets demonstrate that \algo~can outperform state-of-the-art methods by 17\% and 13\%, respectively. An extensive evaluation also proves that \algo~can still perform pretty well even if no sector information is given. Besides, we also present several insights on stock-stock correlation and sector-sector influence by visualizing attention weights.
\end{itemize}
This paper is organized as follows. We first review the relevant studies in Sec.~\ref{sec-related}, followed by presenting the problem statement in Sec.~\ref{sec-prob}. Sec.~\ref{sec-model} describes the technical details of the proposed \algo~model. We report the experimental results in Sec.~\ref{sec-exp}, and conclude this work in Sec.~\ref{sec-conclude}.

\begin{figure*}[!t]
  \centering
  \includegraphics[width=1.0\textwidth]{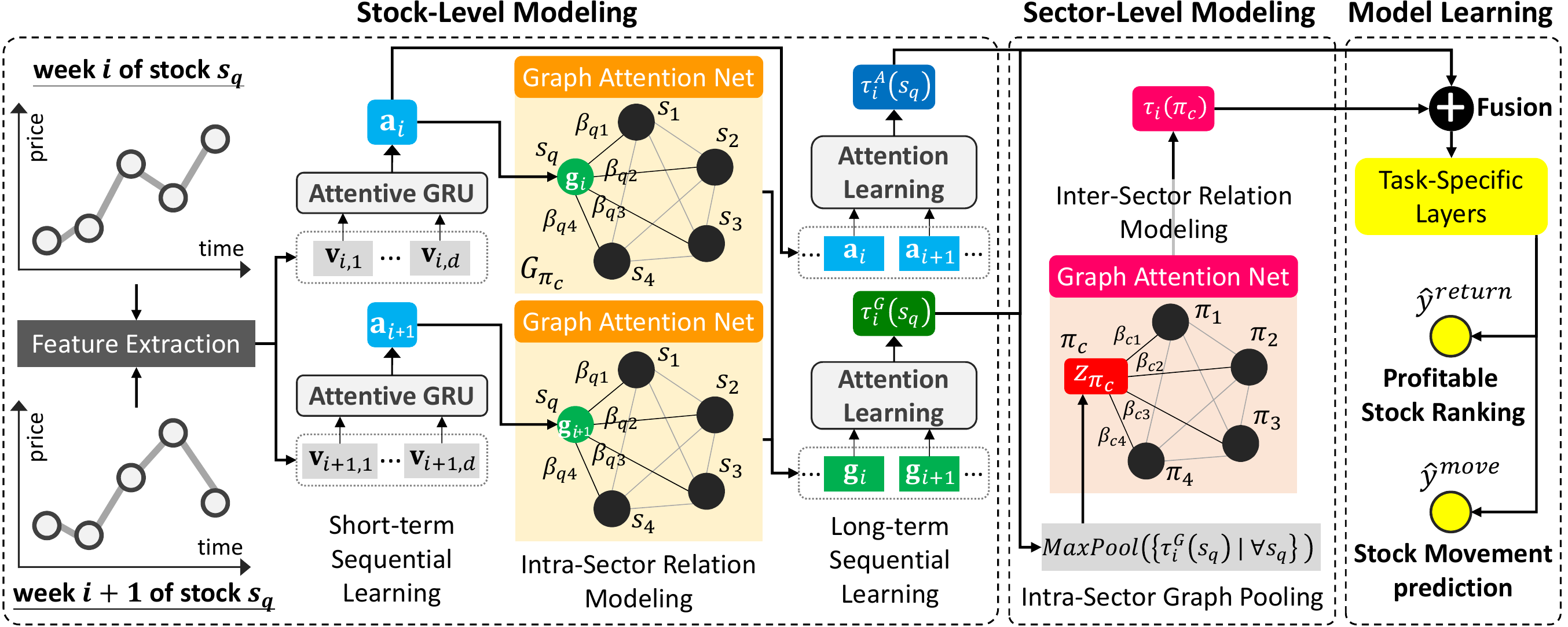}
  \caption{The architecture of our proposed \algo~ model.}
  \label{fig:overall_architecture}
\end{figure*}

\section{Related Work}
\label{sec-related}
Typical methods for stock prediction include ARIMA~\cite{box2015time} and SVR~\cite{cao2003support}. ARIMA considers the linear combination of historical stock prices while SVR treats the stock price at each timestamp independently in the modeling.
By learning the sequential patterns, RNN-based models~\cite{chen2015lstm,baek2018modaugnet}, including LSTM and GRU, improve the prediction performance. SFM~\cite{zhang2017stock} further improves LSTM with memory network and modeling multi-frequency trading patterns.
Attention-based models~\cite{li2018stock} further learn how the next prediction is attended by historical hidden states of stock time series, and produce better results.
Despite that RNN-based models achieve great success in sequential modeling, the performance drops significantly as the length of the sequence increases~\cite{feng2018deepmove}. To capture the long-term dependency of stock movement, FineNet~\cite{tsai2019finenet} utilizes two dilated convolution neural networks to jointly model both long-term and short-term sequential patterns. While most of the existing methods aim at minimizing pointwise loss, a recent RNN-based model, Rank-LSTM~\cite{feng2019temporal}, utilizes pairwise ranking-aware loss, along with pointwise regression loss, to recommend profitable stocks. We will compare the proposed \algo~with Rank-LSTM as it is a state-of-the-art. Ding et al.~\cite{ding2019modeling} propose the extreme value loss (EVL) that enables model to detect extrme stock events.

Some recent advances attempt to model how stocks are affected by one another through learning features from the relationships between stocks. 
Given a graph constructed from a set of pre-defined relationships of investment facts between listed companies, temporal graph convolutional networks~\cite{chen2018incorporating,feng2019temporal,matsunaga2019exploring} are utilized to extract graph-based stock interaction features, and the results are quite promising.
HATS~\cite{kim2019hats} also relies on the pre-defined graph depicting the relations between stocks, but it further learns the embeddings of different types of relationships by considering their hierarchical structure. 
However, we argue that it is unrealistic to presume that the relationships between stocks are always accessible. The relationships are usually hidden due to business concerns. In addition, some influence of price movement between stocks cannot be reflected by their relationships. Our work aims at learning the latent intra-sector and inter-sector relations between stocks. That said, the proposed \algo~does not rely on any pre-defined relationships.

\comment{
Table~\ref{tab:relative_compare} shows the comparison between our \algo~and previous related works. First of all, most of the existing methods~\cite{matsunaga2019exploring,kim2019hats,chen2018incorporating,feng2019temporal} 
utilize graph neural network 
to capture the interaction with intra relationship. However, none research \ct{(which?)} has explored the impact of inter relations between each type. 
We generally solve this issue with learning relational interaction via graph neural network with a learned relation aware embeddings. Secondly, the majority of research~\cite{tsai2019finenet,matsunaga2019exploring} committed to evaluate algorithms by pointwise metrics such as mean square error etc.
Despite the pointwise metrics imply the capability of minimizing distance between prediction and ground-truth,
it's ability for providing investors investment strategy is not measured 
Hence we incorporate a ranking-based loss function to model the relative order by their profitability and therefore evaluate our algorithm with recommendation metrics. Last, while most of the works learn to optimize a single objective, since the accurate prediction on the daily movement is highly relevant to predicting the return ratio of a day, we further enhance our training with the aid of multi-task learning. 

\begin{table}[!t]
\label{tab:relative_compare}
\caption{Contribution comparison of related works and our method.}
\centering
\begin{tabular}{c|c|c|c|c}
\hline
                 & \multicolumn{2}{c|}{Graph Neural Network}         & \multirow{2}{*}{Recommend Stocks} & \multirow{2}{*}{Multi-task learning} \\ \cline{1-3}
                 &   ~~~~Intra~~~~            & Inter                   &                                   &                                      \\ \hline
Tsai, Y.C et al.~\cite{tsai2019finenet} &                           &                           & \checkmark         &                                      \\ \hline
Matsunaga, D. et al.~\cite{matsunaga2019exploring}& \checkmark                           &                           &          &                                      \\ \hline
Kim, R. et al.~\cite{kim2019hats}   & \checkmark &                           &                                   &                                      \\ \hline
Chen, Y. et al.~\cite{chen2018incorporating}  & \checkmark &                           &                                   &                                      \\ \hline
Feng, F. et al.~\cite{feng2019temporal}  & \checkmark &                           & \checkmark         & \checkmark            \\ \hline
Ours             & \checkmark & \checkmark & \checkmark         & \checkmark            \\ \hline
\end{tabular}
\end{table}

}
\section{Problem Statement}
\label{sec-prob}
We are targeting that investors have enough funding but are lack of insights in deciding which of stocks are more worthy to invest among all listed companies. We define the return ratio of stock $s_q$ at the $j$-th day of week $i$, denoted by $R_{ij}^{s_q}$, by considering how much an investor can earn by investing one dollar from the previous day $j-1$, given by:
\begin{equation}
R_{ij}^{s_q} = \frac{p_{ij}^{s_q} - p_{i(j-1)}^{s_q}}{p_{i(j-1)}^{s_q}},
\end{equation}
where $p_{ij}^{s_q}$ is the stock price for stock $s_q$ at the $j$-th day in week $i$.
Let $\mathcal{S} = \{s_1,s_2,...,s_n\}$ denote the universe set of $n$ stocks, and let $\mathbf{v}_{ij}^{s_q}$ denote the feature vector of stock $s_q$ at the $j$-th day of week $i$.
In addition, each stock $s_q$ has its corresponding \textit{sector}, denoted as $\pi_c$, which is an area of the economy that businesses share the same or a related product or service. The concept of sector can be also thought of as an industry or market that shares common operating characteristics. A sector can contain multiple stocks. Given the feature matrix $\mathbf{D}_{i}^{s_q} = \{\mathbf{v}^{s_q}_{i(j-d)}, \mathbf{v}^{s_q}_{i(j-d+1)},...,\mathbf{v}^{s_q}_{i(j-1)} \}$ derived from the $(j-d)$-th to $(j-1)$-th day in week $i$, in which $d$ is the number of past days to construct the feature matrix,
our goal is to first predict the return ratio $R_{(i+1)1}^{s_q}$ of every stock $s_q\in\mathcal{S}$, then to accordingly recommend a list of top-$K$ stocks for the first day of week $i+1$.


\section{The Proposed \algo~Model}
\label{sec-model}
In this section, we first explain the model architecture of our \algo, which is demonstrated in Figure~\ref{fig:overall_architecture}. 
The proposed \algo~consists of three main components: (1) stock-level modeling, (2) sector-level modeling, and (3) model training. First, in stock-level modeling, we begin with extracting several sequential features from the price time series of each stock $s_q$ at every week $i$. The features are utilized for short-term sequential learning through an attentive gated recurrent unit (Attentive GRU). The short-term feature representation vector $\mathbf{a}_i$ can be generated. Then in intra-sector relation modeling, we aim to model the latent relationships between stocks that belong to the same sector $\pi_c$. A fully-connected graph is created to depict the hidden links among same-sector stocks, and a graph attention network is applied to learn and produce the graph-based feature representation $\mathbf{g}_i$ for stock $s_q$. The last step of the first component is long-term sequential learning, which generates the aggregated embedding vectors over past weeks using attention learning. Two vectors $\tau^A_i(s_q)$ and $\tau^G_i(s_q)$ are produced based on short-term embeddings $\mathbf{a}_i$ and graph-based embeddings $\mathbf{g}_i$, respectively. Second, in sector-level modeling, we create a fully-connected graph to represent the potential interactions between sectors. An intra-sector graph pooling is used to compute each sector $\pi_c$'s initial embedding $\mathbf{z}_{\pi_c}$ over same-sector stocks' embeddings $\tau^G_i(s_q)$. Then in inter-sector relation modeling, we apply a graph attention network to a sector-wise fully-connected graph, and produce the inter-sector embeddings $\tau_i(\pi_c)$ that capture sector interactions. Third, we fuse the derived embeddings $\tau^A_i(s_q)$, $\tau^G_i(s_q)$, and $\tau_i(\pi_c)$ via concatenation, and use the fused embedding to perform multi-task training. One task is profitable stock ranking, and the other is stock movement prediction.

\subsection{Stock-level Modeling}
\label{sec:stocklevel}
\textbf{Feature Extraction.} We define the initial features of each stock at one day. We have basic daily features of each stock at day $j$, including opening $open_j$ and closing prices $close_j$, the highest $high_j$ and lowest $low_j$ prices, the adjusted closing prices $adjclose_j$, and the return ratio. We also define hand-craft features, including price-ratio features:
\begin{equation}
\mathcal{F}_{\mu} = \frac{\mu_j}{close_j}-1
\end{equation}
where $\mu\in\{open, high, low\}$, 
and moving-average features:
\begin{equation}
\mathcal{F}_{\varphi} = \frac{\sum_{j=0}^{\varphi-1} adjclose_j}{\varphi} / (adjclose_j-1),
\end{equation}
where $\varphi\in\{5, 10, 15, 20, 25, 30\}$.
We concatenate these feature values to be the initial feature vectors $\mathbf{v}_{ij}^{s_q}$ for stock $s_q$ at the $j$-th day in week $i$. 


\textbf{Short-term Sequential Learning.}
We exploit recurrent neural networks to learn short-term sequential features. The sequence of initial feature vectors $\mathbf{v}_{ij}^{s_q}$ for days within a week is used as the input. 
Gated Recurrent Unit (GRU)~\cite{cho2014learning} is adopted, 
and its last hidden-state vector $\mathbf{h}_{ij}^{s_q}$ is week $i$'s output embedding, given by:
\begin{equation}
\mathbf{h}_{ij}^{s_q} = GRU(\mathbf{v}_{ij}^{s_q},\mathbf{h}_{i(j-1)}^{s_q}),
\end{equation}
where
$\mathbf{h}_{i(j-1)}^{s_q}$ is the hidden-state vector of day $j-1$. 
The feed-forward neural network-based attention mechanism~\cite{luong2015effective}
is used to learn dynamic weights depicting the importance of each day, and aggregate day-wise hidden-state vectors to better encode week $i$'s sequential patterns. 
Let $H_i = \{\mathbf{h}_{i1}^{s_q},\mathbf{h}_{i2}^{s_q},...,\mathbf{h}_{id}^{s_q}\}$ be the input of attention selector. We generate the attentive representation vector $\mathbf{a}_i^{s_q}$ of stock $s_q$ at week $i$ by:
\begin{equation}
\begin{aligned}
\mathbf{a}_i^{s_q} = Attention(H_i) = \sum_{j} \alpha_j^{s_q} \mathbf{h}_{ij}^{s_q},
\end{aligned}
\end{equation}
\begin{equation}
\alpha_j^{s_q} = \sigma(tanh(\mathbf{W}_0 \mathbf{h}_{ij}^{s_q}),
\end{equation}
where $\mathbf{W}_0$ is the matrix of learnable parameters, and $\sigma$ is the softmax function. Note that here $j$ refers to each day within week $i$.


\textbf{Intra-sector Relation Modeling.}
Here we aim at modeling the latent relationships between stocks that belong to the same sector. Instead of presuming any explicit relations between stocks are given, we allow that two stocks could influence one another. Hence, for each sector $\pi_c$, we create a fully-connected graph, in which any two stocks belonging to $\pi_c$ are directly connected, as demonstrated by links that connect same-color nodes $s_q$ in Figure~\ref{fig:Toy_example}. The intra-sector influence will be learned through graph neural networks, and is encoded in the output embedding vector of each stock.
Let $G_{\pi_c}=(M_{\pi_c},E_{\pi_c})$ denote the \textit{intra-sector graph} for sector $\pi_c$, where $M_{\pi_c}$ is the set of listed companies belonging to $\pi_c$, and $E_{\pi_c}$ is the set of edges between two stocks $s_q,s_r$, where $s_q\in M_{\pi_c}$ and $s_r\in M_{\pi_c}$. That said, for each stock $s_q\in M_{\pi_c}$, we create an edge that connects $s_q$ to every of other stocks $s_r \in M_{\pi_c}$, where $s_r \neq s_q$.
The initial feature vector of each node $s_q$ in $G_{\pi_c}$ is the embedding $\mathbf{a}_i^{s_q}$ that encodes sequential patterns in stock time series within a week.
Since the relationship strength between every pair of stocks could vary, Graph Attention Network (GAT)~\cite{velivckovic2017graph} is adopted to be the graph neural network model. With GAT, each stock node can use various learned contributions to absorb information from other stock nodes. In other words, GAT is considered to simulate how stocks are interacted and influenced with one another based on their time series.

The graph attention network is to learn attention weights between nodes, and utilizes the weights to aggregate information from the neighbors of each node. It can be mathematically represented as follows:
\begin{equation}
    GAT(G_{\pi_c};s_q) = ReLU\left(\sum_{s_n \in \Gamma(s_q)} \beta_{qn}\mathbf{W}_1\mathbf{a}_i^{s_n}\right) ,
\end{equation}
where $\beta_{qn}$ denotes the attention weight from stock $s_n$ to stock $s_q$, $\Gamma(s_q)$ returns the set of neighbors for node $s_q$ in $G_{\pi_c}$, and $\mathbf{W}_1$ is a learnable weight matrix. The attention weights $\beta_{qn}$ can be derived by:
\begin{equation}
    \beta_{qn} = \frac{\exp\left(LeakyReLU\left(\mathbf{r}^\top [\mathbf{W}_2 \mathbf{a}_i^{s_q} \parallel \mathbf{W}_2 \mathbf{a}_i^{s_n}]\right)\right)}{\sum_{s_n \in \Gamma(s_q)} \exp\left(LeakyReLU\left(\mathbf{r}^\top [\mathbf{W}_2 \mathbf{a}_i^{s_q} \parallel \mathbf{W}_2 \mathbf{a}_i^{s_n}]\right)\right)},
\end{equation}
where $\mathbf{r}$ 
is the learnable vector that projects the embedding into a scalar, $\mathbf{\parallel}$ denotes the concatenation operation, and $\mathbf{W}_2$ is a learnable weight matrix. With the aid of graph attention network, we can generate the graph-based representation $\mathbf{g}_{i}^{s_q} = GAT(G_{\pi_c};s_q)$ that encodes intra-sector relations i.e., how stock $s_q$ is interacted and influenced by other stocks in week $i$. 
That is, $\mathbf{g}_{i}^{s_q}$ can be seen as the combination of other stocks weighted by graph attention that considers the similarity between stocks.



\textbf{Long-term Sequential Learning.}
Since the stock price could be affected by both short-term and long-term movements~\cite{tsai2019finenet}, we aim to aggregate the derived short-term (i.e., week-level) embedding vectors to obtain long-term sequential features.
Here we consider two kinds of temporal information. One is the embedding $\mathbf{a}^{s_q}_{i}$ that encodes the primitive long-term sequential features. The other is the short-term embedding $\mathbf{g}^{s_q}_{i}$ that incorporating the learning of intra-sector relations. Assume that the past $t$ weeks are used to learn long-term features of a stock. We accordingly have two sequences of short-term embeddings: 
\begin{equation}
\begin{aligned}
U_i^G(s_q) &= \{\mathbf{g}^{s_q}_{i-t}, \mathbf{g}^{s_q}_{i-t+1}, ..., \mathbf{g}^{s_q}_{i-1}\},\\
U_i^A(s_q) &= \{\mathbf{a}^{s_q}_{i-t}, \mathbf{a}^{s_q}_{i-t+1}, ..., \mathbf{a}^{s_q}_{i-1}\},
\end{aligned}
\end{equation}
which are obtainable from week $i-t$ to week $i-1$.
Here we separately apply attentive GRU (described before) to $U_i^G(s_q)$ and $U_i^A(s_q)$ to generate two long-term embedding vectors, denoted by $\tau_i^G(s_q)$ and $\tau_i^A(s_q)$, respectively. The corresponding process can be represented by:
\begin{equation}
\begin{aligned}
    \tau_i^G(s_q) &= Attention\left(U_i^G(s_q)\right), \\
    \tau_i^A(s_q) &= Attention\left(U_i^A(s_q)\right).
\end{aligned}
\end{equation}
The long-term embedding vectors $\tau_i^G(s_q)$ and $\tau_i^A(s_q)$ are produced by not only modeling the week-wise sequential features, but also being effectively combined through learnable attention weights.
\subsection{Sector-level Modeling}
\label{sec:sectorlevel}
This section aims at learning how different sectors are influenced and interacted with one another by modeling their latent relations. Given the intra-sector embeddings $\tau_i^G(s_q)$ of stocks belonging to sector $\pi_c$ at week $i$, we first need to generate the initial sector embeddings by intra-sector graph pooling. Then we perform inter-sector relation modeling to learn sector-sector interactions.

\textbf{Intra-sector Graph Pooling.} We first generate a sector embedding from stocks that belong to the sector $\pi_c$. 
Graph pooling methods~\cite{lee2019self,cho2014finding} can be used to obtain an unified vector from graph $G_{\pi_c}$, 
based on the long-term embeddings $\tau_i^G(s_q)$ of stocks $s_q\in M_{\pi_c}$.
Given a sector specific graph $G_{\pi_c}$, we use the element-wise max-pooling operation to generate an embedding $\mathbf{z}_{\pi_c}$ 
that represents sector $\pi_c$ by:
\begin{equation}
\mathbf{z}_{\pi_c} = MaxPool\left(\{ \tau_i^G(s_q)\,\,|\,\, \forall s_q \in M_{\pi_c} \}\right).
\end{equation}
The operation $MaxPool$ is the element-wise max pooling that generates a vector from a set of vectors, given by: $MaxPool(X)=[\max(\{x_1|\forall x\in X\}), \max(\{x_2|\forall x\in X\}), ..., \max(\{x_\epsilon|\forall x\in X\})]$, where $x$ is the $\epsilon$-dimensional vector, $x_k$ is the $k$-th element in vector $x$, $X$ is a set of $\epsilon$-dimensional vectors, and the operation $\max$ takes the maximum in a set of values.
We choose element-wise max pooling instead of other pooling methods
due to its simplicity without any learnable parameters.
Eventually, a set of sector embeddings $Z_{\pi} = \{\mathbf{z}_{\pi_1}, \mathbf{z}_{\pi_2},...,\mathbf{z}_{\pi_c} \}$ can be obtained, where $c$ is the number of sectors. 

\textbf{Inter-sector Relation Modeling.}
Since the interactions between sectors are hidden and dynamic in some latent relations, we construct a fully-connected graph $G_\pi=(Z_{\pi},E_\pi)$, where $Z_{\pi}$ is the set of all sectors with their embeddings (i.e., each sector is considered as a node)
, and $E_\pi$ is the set of edges that directly connect every pair of sector nodes in graph $G_\pi$.
To model the high-order interactions through the latent relations between sectors, we again adopt graph attention network. The sector embeddings $\mathbf{z}_{\pi_c}$ are used to initialize the vectors of sector nodes in $G_\pi$. The inter-sector embeddings $\tau_i(\pi_c)$ of sector $\pi_c$ generated by GAT is given by:
\begin{equation}
\tau_i(\pi_c) = GAT(G_\pi,\pi_c).
\end{equation}
The derived embeddings $\tau_i(\pi_c)$ encodes how sectors are influenced by each other with various attention weights in either direct or indirect manner.

\subsection{Model Learning}
\label{sec:modellearn}
\textbf{Embedding Fusion.}
The proposed \algo~incorporates a variety of features to predict the return ratio of stocks. The derived feature vectors, including the primitive short-term embeddings $\tau_i^A(s_q)$, intra-sector embeddings $\tau_i^G(s_q)$, and inter-sector embeddings $\tau_i(\pi_c)$, are used.
We combine these features via an embedding fusion layer to obtain the final feature vector $\tau_i^F(s_q)$, given by:
\begin{equation}
\label{finalfv}
\tau_i^F(s_q) = ReLU\left(\left[\tau_i^G(s_q) \parallel \tau_i^A(s_q) \parallel \tau_i(\pi_c)\right]\mathbf{W}_f\right),
\end{equation}
where stock $s_q$ belongs to sector $\pi_c$, $\mathbf{W}_f$
is the learnable weight matrix, and ReLU is the activation function. In other words, the past $t$ weeks, i.e., from week $i-t$ to week $i-1$, is used to produce the final embedding vector $\tau_i^F(s_q)$ at the first day of the $i$-th week, and to predict the corresponding daily return ratio.

\textbf{Multi-Task Learning.}
Recommending the most profitable stocks can be divided into two correlated parts: ranking stocks based on their predicted return ratios, and finding future stocks with positive movements (i.e., stocks that go up). Therefore, rather than adopting point-wise loss (e.g., mean squared error) that cannot reflect the profitability of stocks, we resort to jointly optimize the ranking of stocks based on predicted return ratio and the movements of stocks (i.e., binary labels of up and down). That said, we aim at exploiting the concept of multi-task learning in the optimization of \algo. In predicting the ranking, we utilize a pairwise ranking-aware loss~\cite{feng2019temporal}, which encourages the ranking order of a stock pair based on predicted return ratio to have the same order as the ranking based on their ground-truth return ratio. In predicting the movement, the cross-entropy loss is employed. The predictions of return ratio and movement for stock $s_q$, denoted by $\hat{y}_i^{return}(s_q)$ and $\hat{y}_i^{move}(s_q)$, can be performed by their respective task-specific layers, given by:
\begin{equation}
\begin{aligned}
\label{eq:task_prediction}
\hat{y}_i^{return}(s_q) &= \mathbf{e}_1^\top \tau_i^F(s_q) + \mathbf{b}_1,\\
\hat{y}_i^{move}(s_q) &= \phi\left(\mathbf{e}_2^\top \tau_i^F(s_q) + \mathbf{b}_2\right),
\end{aligned}
\end{equation}
where $\mathbf{e}_1, \mathbf{e}_2 \in \mathbb{R}^d$ denote the hidden vectors of task-specific layers that project $\tau_i^F(s_q)$ into the prediction results of return ratio and binary movement,
respectively, $\phi$ is the sigmoid function, and $\mathbf{b}_1$ and $\mathbf{b}_2$ are bias terms. $\hat{y}_i^{return}(s_q)$ is the predicted value of return ratio at week $i$, and $\hat{y}_i^{move}(s_q)$ is the predicted probability of the return ratio at week $i$ being positive. Note that the subscript $i$ in Equation~\ref{finalfv} and Equation~\ref{eq:task_prediction} is used to denote that ``we are generating the stock's embedding based on the past $t$ weeks, and are predicting the return ratio at the first day of the $i$-th week.'' To generate the daily prediction results, we utilize a daily sliding window to have data instances. The details are described in Section~\ref{sec:expsetting}.

The proposed \algo~is to optimize a multi-task objective that simultaneously predicts the ranking of return ratio and the movement of stocks. 
The final loss function, denoted by $\mathcal{L}_{\algo}$, is given by:
\begin{equation}
\begin{aligned}
    \mathcal{L}_{\algo} = (1-\delta) \mathcal{L}_{rank} + \delta \mathcal{L}_{move} + \lambda \| \Theta \|^2,
\end{aligned}
\label{eq:loss}
\end{equation}
where $\mathcal{L}_{rank}$ is the pairwise ranking loss in terms of return ratio, and $\mathcal{L}_{move}$ is the cross-entropy loss for binary movement classification. $\Theta$ depicts all learnable weights, and $\lambda$ is the regularization hyperparameter to prevent overfitting. Such two loss functions are given by: 
\begin{equation}
\begin{aligned}
    \mathcal{L}_{rank} &= \sum_i \sum_{s_q} \sum_{s_k} max\left(0,-\hat{\Delta} \times \Delta \right), \\
    \text{where }\hat{\Delta} &= \left(\hat{y}_i^{return}(s_q) - \hat{y}_i^{return}(s_k)\right), \\
    \Delta &= \left(y_i^{return}(s_q) - y_i^{return}(s_k)\right), \\ \\
    \mathcal{L}_{move} &= - \sum_i \sum_{s_q} \,\,\, y^{move}_{i} \log\left(\hat{y}_i^{move}(s_q)\right) \\
    &+ \left(1-y^{move}_{i}\right) \log\left(1-\hat{y}_i^{move}(s_q)\right),
\end{aligned}
\end{equation}
where $y_i^{return}(s_q)$ is the ground-truth return ratio at week $i$, and $y_i^{move}(s_q)$ 
is the groud-truth binary label ($1$ is assigned if the ground-truth return ratio at week $i$
is positive and $0$ for negative). $\delta$ is a hyperparameter that determines the balance between two prediction tasks.

\begin{table*}[!t]
\centering
\caption{Statistics of two stock datasets.}
\label{tab:dataset}
\resizebox{0.9\textwidth}{!}{%
\begin{tabular}{c|c|c|c|c|c}
\hline
Market       & \# Stocks & \# Sectors & \begin{tabular}[c]{@{}c@{}} \# Training Days\end{tabular} & \begin{tabular}[c]{@{}c@{}} \# Validation Days\end{tabular} & \begin{tabular}[c]{@{}c@{}} \# Testing Days\end{tabular} \\ \hline
Taiwan Stock & 100  & 5  & 579   & 193   & 193   \\ \hline
S\&P 50 & 424    & 9     & 579   & 193   & 193   \\ \hline
NASDAQ & 1026    & 112     & 756   & 252   & 237   \\ \hline
\end{tabular}
}
\end{table*}

\section{Evaluation}
\label{sec-exp}
We conduct a series of experiments to answer the following five evaluation questions. 
\begin{itemize}
\item \textbf{EQ1:} Can \algo~outperform state-of-the-art (SOTA) models on top-$K$ stock recommendation?
\item \textbf{EQ2:} Will \algo~be still able to have promising performance if no sector information is given?
\item \textbf{EQ3:} Does each component of \algo~effectively contribute to the recommendation performance?
\item \textbf{EQ4:} How do various hyperparameters affect the recommendation performance of \algo?
\item \textbf{EQ5:} What can be captured by intra-sector and inter-sector graph attention weights in \algo?
\end{itemize}

\subsection{Evaluation Settings}
\label{sec:expsetting}
\textbf{Datasets.}
We employ three real-world financial dataset: Taiwan Stock~\footnote{\url{https://www.twse.com.tw/en/page/listed/listed\_company/new\_listing.html}}, S\&P 500~\footnote{\url{https://datahub.io/core/s-and-p-500}} and NASDAQ~\footnote{\url{https://github.com/fulifeng/Temporal_Relational_Stock_Ranking}}.
Each dataset contains daily stock prices and sector information of every listed company. 
For Taiwan stock and S\&P 5 dataset, we consider 60\% for training (579 days), 20\% for validation (193 days), and 20\% for testing (193 days). 
Every consecutive $16$ trading days is treated as a data instance, consisting of: $3$ weeks (five days in a week) as training data, 
and the $16$-th day (e.g., the first day of the fourth week) as the prediction target.
That said, to predict the daily return ratio, we utilize a sliding window with $15$ days to compile the data, and each of its next day (i.e., the $16$-th day) is used for prediction.
For NASDAQ, we follow the setting in RankLSTM~\cite{feng2019temporal}.
By filtering out those stocks whose data instances cannot satisfy the training-validation-test setting, we have the data statistics summarized in Table~\ref{tab:dataset}.

\textbf{Evaluation Metrics.}
The list of ground-truth return ratio of all stocks $S$ in $j$-th day of week $i$ is denoted as $R^S_{ij} = \{R_{ij}^{s_1},R_{ij}^{s_2}...,R_{ij}^{s_n}\}$, in which $R_{ij}^{s_q}$ is the ground-truth return ratio of stock $s_q$. The predicted return ratio is denoted by $\hat{R}^S_{ij}$ and $\hat{R}^{s_q}_{ij}$ for all stocks $S$ and a single stock $s_q$, respectively.
We rank stocks based on their return ratio. Stocks with higher return ratio are ranked at top positions.
The lists of predicted and ground-truth top-$K$ stocks are denoted as $L@K(R^S_{ij})$ and $L@K(\hat{R}^S_{ij})$, respectively.
Since our goal is to recommend the most profitable stocks, metrics on error measures, such as MAE and RMSE, are not adopted.
We employ three evaluation metrics that are widely used in recommender systems. 
\begin{itemize}
\item \textbf{Mean Reciprocal Rank (MRR@K)}:
\begin{equation}
MRR@K = \frac{1}{K}\sum_{s_q\in L@K(\hat{R}^S_{ij})}\frac{1}{rank(R^{s_q}_{ij})},
\end{equation}
where $rank(R^{s_q}_{ij})$ returns the ground-truth rank of stock $s_q$ at the $j$-th day in week $i$.
\item \textbf{Precision@K}:
\begin{equation}
Precision@K = \frac{| L@K(\hat{R}^S_{ij}) \cap L@K(R^S_{ij}) |}{K}
\end{equation}
\item \textbf{Accuracy (ACC)}: ACC is the number of correct predictions of binary movement divided by the number of testing instances.
\end{itemize}
In all of the metrics, higher scores indicate better performance. The experiments are executed based on the aforementioned setting of data splitting that relies on the sliding window. The ``\# Testing Days $\times$ \# Stocks'' results are generated for each dataset. The average scores of 10 runs produced from testing data are reported. The same evaluation procedure is applied to both \algo~and all competing methods.

\textbf{Competing Methods.}
We compare the proposed \algo~with following five methods. In each method, we first generate the predicted results of return ratio, then accordingly rank stocks. 
\begin{enumerate}
\item \textbf{MLP}~\cite{tang2015extreme}: multi-layer perceptron using two hidden layers with 32 and 8 dimensions.
\item \textbf{GRU}~\cite{cho2014learning}: a compact RNN-based model with a 32-dimensional GRU layer to learn sequential features from time series.
\item \textbf{GRU+Att}~\cite{dhingra2016gated}: combining one 32-dimensional GRU layer with an attention layer that gives various contribution weights to each time step.
\item \textbf{FineNet}~\cite{tsai2019finenet}: a state-of-the-art joint convolutional and recurrent neural network (with one 32-dim and one 16-dim convolution layers) that is effective in capturing short-term and long-term patterns of stock time series.
\item \textbf{RankLSTM}~\cite{feng2019temporal}: a state-of-the-art graph neural network-based method that utilizes temporal graph convolution with 16-dim embeddings to model time series of stocks, and incorporating pairwise ranking loss for recommending stocks. The balancing hyperparameter $\alpha$ in RankLSTM is searched: $\alpha\in\{0.01,0.1,1,10\}$.
\end{enumerate}


\textbf{Settings of \algo.}
The dimensions for hidden layers of GRU and GAT are all set to $16$. The learning rate is searched in $\{0.0005, 0.001, 0.005\}$. The batch size is set to $128$, and the balancing hyperparameter is $\delta=0.01$. The regularization parameter is $\lambda=0.0001$. We optimize all the models using the Adam optimizer~\cite{kingma2014adam}. All experiments are conducted with PyTorch\footnote{https://pytorch.org/} and PyTorch Geometric\footnote{https://pytorch-geometric.readthedocs.io/en/latest/}, which is based on Python programming language, running on GPU machines (Nvidia GeForce GTX 1080 Ti). The implementation can be found in \url{https://github.com/Roytsai27/Financial-GraphAttention}.

\subsection{Experimental Results}
\begin{table*}[!t]
\centering
\caption{Main experimental results by varying top-$K$, $K\in\{5,10,20\}$.
}
\label{tab:main_result}
\resizebox{0.85\textwidth}{!}{%
\begin{tabular}{cccccccc}
\hline
\multicolumn{8}{c}{Taiwan Stock}                                                                                                                                                                                  \\ \hline
\multicolumn{1}{c|}{}           & \multicolumn{2}{c|}{$K=5$}               & \multicolumn{2}{c|}{$K=10$}              & \multicolumn{2}{c|}{$K=20$}                                 & Overall                     \\ \cline{2-8} 
\multicolumn{1}{c|}{Model}      & MRR     & \multicolumn{1}{c|}{Precision} & MRR     & \multicolumn{1}{c|}{Precision} & MRR                        & \multicolumn{1}{c|}{Precision} & ACC                         \\ \hline
\multicolumn{1}{c|}{MLP}        & 0.2842  & \multicolumn{1}{c|}{0.0500}    & 0.5753  & \multicolumn{1}{c|}{0.1022}    & 1.0114                     & \multicolumn{1}{c|}{0.1992}    & 0.4514                      \\
\multicolumn{1}{c|}{GRU}        & 0.3115  & \multicolumn{1}{c|}{0.0622}    & 0.6222  & \multicolumn{1}{c|}{0.1272}    & 1.0639                     & \multicolumn{1}{c|}{0.2053}    & 0.4812                      \\
\multicolumn{1}{c|}{GRU+Att}    & 0.3435  & \multicolumn{1}{c|}{0.0811}    & 0.6736  & \multicolumn{1}{c|}{0.1417}    & 1.1779                     & \multicolumn{1}{c|}{0.2131}    & 0.4948                      \\
\multicolumn{1}{c|}{FineNet}    & 0.3742  & \multicolumn{1}{c|}{0.0867}    & 0.7002  & \multicolumn{1}{c|}{0.1572}    & 1.2500                     & \multicolumn{1}{c|}{0.2206}    & 0.5295                      \\
\multicolumn{1}{l|}{RankLSTM} & 0.3962  & \multicolumn{1}{c|}{0.1011}    & 0.7838  & \multicolumn{1}{c|}{0.1717}    & \multicolumn{1}{l}{1.3298} & \multicolumn{1}{c|}{0.2456}    & 0.5539                      \\ \hline
\multicolumn{1}{c|}{\algo}     & \textbf{0.4391}  & \multicolumn{1}{c|}{\textbf{0.1133}}    & \textbf{0.8479}  & \multicolumn{1}{c|}{\textbf{0.2022}}    & \textbf{1.4106}                     & \multicolumn{1}{c|}{\textbf{0.2756}}    & \textbf{0.5682}                      \\ \hline
\multicolumn{1}{c|}{Improv.}    & 10.83\% & \multicolumn{1}{c|}{12.08\%}   & 8.18\%  & \multicolumn{1}{c|}{17.76\%}   & 6.08\%                     & \multicolumn{1}{c|}{12.21\%}   & 2.58\%                      \\ \hline
\multicolumn{8}{c}{S\&P 500}                                                                                                                                                                                       \\ \hline
\multicolumn{1}{c|}{}           & \multicolumn{2}{c|}{$K=5$}               & \multicolumn{2}{c|}{$K=10$}              & \multicolumn{2}{c|}{$K=20$}                                 & \multicolumn{1}{l}{Overall} \\ \cline{2-8} 
\multicolumn{1}{c|}{Model}      & MRR     & \multicolumn{1}{c|}{Precision} & MRR     & \multicolumn{1}{c|}{Precision} & MRR                        & \multicolumn{1}{c|}{Precision} & ACC                         \\ \hline
\multicolumn{1}{c|}{MLP}        & 0.0844  & \multicolumn{1}{c|}{0.0172}    & 0.1828  & \multicolumn{1}{c|}{0.0215}    & 0.3886                     & \multicolumn{1}{c|}{0.0594}    & 0.535                       \\
\multicolumn{1}{c|}{GRU}        & 0.1158  & \multicolumn{1}{c|}{0.0266}    & 0.2229  & \multicolumn{1}{c|}{0.0366}    & 0.4390                     & \multicolumn{1}{c|}{0.0685}    & 0.5342                      \\
\multicolumn{1}{c|}{GRU+Att}    & 0.1321  & \multicolumn{1}{c|}{0.0301}    & 0.2513  & \multicolumn{1}{c|}{0.0516}    & 0.4813                     & \multicolumn{1}{c|}{0.0849}    & 0.5353                      \\
\multicolumn{1}{c|}{FineNet}    & 0.1502  & \multicolumn{1}{c|}{0.0387}    & 0.2965  & \multicolumn{1}{c|}{0.0602}    & 0.5392                     & \multicolumn{1}{c|}{0.0909}    & 0.539                       \\
\multicolumn{1}{l|}{RankLSTM} & 0.1736  & \multicolumn{1}{c|}{0.0398}    & 0.3034  & \multicolumn{1}{c|}{0.0597}    & 0.5411                     & \multicolumn{1}{c|}{0.0911}    & 0.5411                      \\ \hline
\multicolumn{1}{c|}{\algo}     & \textbf{0.1974}  & \multicolumn{1}{c|}{\textbf{0.0419}}    & \textbf{0.3357}  & \multicolumn{1}{c|}{\textbf{0.0677}}    & \textbf{0.5687}                     & \multicolumn{1}{c|}{\textbf{0.0989}}    & \textbf{0.5425}                      \\ \hline
\multicolumn{1}{c|}{Improv.}    & 13.71\% & \multicolumn{1}{c|}{5.28\%}    & 10.65\% & \multicolumn{1}{c|}{12.46\%}   & 5.10\%                     & \multicolumn{1}{c|}{8.56\%}    & 0.26\%                      \\ \hline

\multicolumn{8}{c}{NASDAQ}                                                                                                                                                                                       \\ \hline
\multicolumn{1}{c|}{}           & \multicolumn{2}{c|}{$K=5$}               & \multicolumn{2}{c|}{$K=10$}              & \multicolumn{2}{c|}{$K=20$}                                 & \multicolumn{1}{l}{Overall} \\ \cline{2-8} 
\multicolumn{1}{c|}{Model}      & MRR     & \multicolumn{1}{c|}{Precision} & MRR     & \multicolumn{1}{c|}{Precision} & MRR                        & \multicolumn{1}{c|}{Precision} & ACC                         \\ \hline
\multicolumn{1}{c|}{MLP}        & 6.13e-3  & \multicolumn{1}{c|}{1.12e-3}    & 8.95e-3  & \multicolumn{1}{c|}{2.01e-3}    & 2.10e-2                     & \multicolumn{1}{c|}{3.56e-3}    & 0.1374                       \\
\multicolumn{1}{c|}{GRU}        & 9.31e-3  & \multicolumn{1}{c|}{3.85e-3}    & 2.14e-2  & \multicolumn{1}{c|}{4.89e-3}    & 3.76e-2                    & \multicolumn{1}{c|}{6.03e-3}    & 0.1758                      \\
\multicolumn{1}{c|}{GRU+Att}    & 9.28e-3  & \multicolumn{1}{c|}{3.94e-3}    & 2.56e-2  & \multicolumn{1}{c|}{5.04e-3}    & 3.45e-2                     & \multicolumn{1}{c|}{5.85e-3}    & 0.1539                      \\
\multicolumn{1}{c|}{FineNet}    & 1.34e-2  & \multicolumn{1}{c|}{4.18e-3}    & 2.83e-2  & \multicolumn{1}{c|}{6.14e-3}    & 3.85e-2                    & \multicolumn{1}{c|}{6.45e-3}    & 0.1905                       \\
\multicolumn{1}{l|}{RankLSTM} & 1.81e-2  & \multicolumn{1}{c|}{4.35e-3}    & \textbf{3.43e-2}  & \multicolumn{1}{c|}{\textbf{6.78e-3}}    &    4.16e-2                  & \multicolumn{1}{c|}{7.04e-3}    & 0.2318                      \\ \hline
\multicolumn{1}{c|}{\algo}     & \textbf{2.03e-2}  & \multicolumn{1}{c|}{\textbf{4.63e-3}}    & 3.01e-2  & \multicolumn{1}{c|}{6.24e-3}    & \textbf{4.57e-2}                     & \multicolumn{1}{c|}{\textbf{7.15e-3}}    & \textbf{0.2579}                      \\ \hline
\multicolumn{1}{c|}{Improv.}    & 12.15\% & \multicolumn{1}{c|}{6.43\%}    & -12.22\% & \multicolumn{1}{c|}{-7.96\%}   & 9.85\%                     & \multicolumn{1}{c|}{1.56\%}    & 11.25\%                      \\ \hline
\end{tabular}
}
\end{table*}

\textbf{Main Results.} Table~\ref{tab:main_result} displays the main experimental results. 
We can find that the proposed \algo~significantly outperforms all of the competing methods among three datasets, especially on the ranking evaluation in terms of MRR and Precision.
When recommending fewer $K$ stocks, \algo~leads to particularly good performance. For example, \algo~outperforms RankLSTM by $13.71\%$ and $12.15\%$ in terms of MRR when $K=5$ on S\&P 500 and NASDAQ datasets, respectively. 
\algo~also generates higher precision scores than RankLSTM by $17.76\%$ when $K=10$ on Taiwan Stock data. Although the performance of \algo~on NASDAQ with $K=10$ is not satisfying (i.e., worse than RankLSTM), the proposed \algo~consistently leads to significant top-$5$ ($K=5$) recommendation performance improvement against RankLSTM across three datasets. The average improvements are averagely $12.23\%$ and $7.93\%$ on MRR and Precision, respectively. While accurately recommending items at the top positions would better benefit user decision because people tend to believe top recommended ones with smaller $K$ values~\cite{topkkdd20,recsuv19}, the consistent and significant performance boosting of our \algo~with $K=5$ is confirmed to achieve the most useful stock recommendation for users.
Such results imply that \algo~is able to accurately recommend the most profitable stocks. The promising results of \algo, comparing to the state-of-the-art models, FineNet and RankLSTM, reflect the effectiveness of modeling the interactions between stocks, between sectors, and between stocks and sectors under the proposed architecture of graph-aware hierarchical information passing. In other words, the main reason that the competing methods cannot be competitive as \algo~is not well leveraging and learning the stock-stock, stock-sector, and sector-sector interactions. Moreover, the results that \algo~outperforms RankLSTM can verify that learning the latent relationships between stocks can better depict the influence between stocks than learning based on pre-defined relations between stocks. Although our \algo~cannot have the same significant improvement on the prediction of stock movement in terms of ACC, it still leads to higher accuracy values than the competing methods. That said, a minor weakness of \algo~lies in cannot further produce significant performance improvement on stock movement prediction. 

\begin{figure*}[!t]
  \centering
  \includegraphics[width=0.9\textwidth]{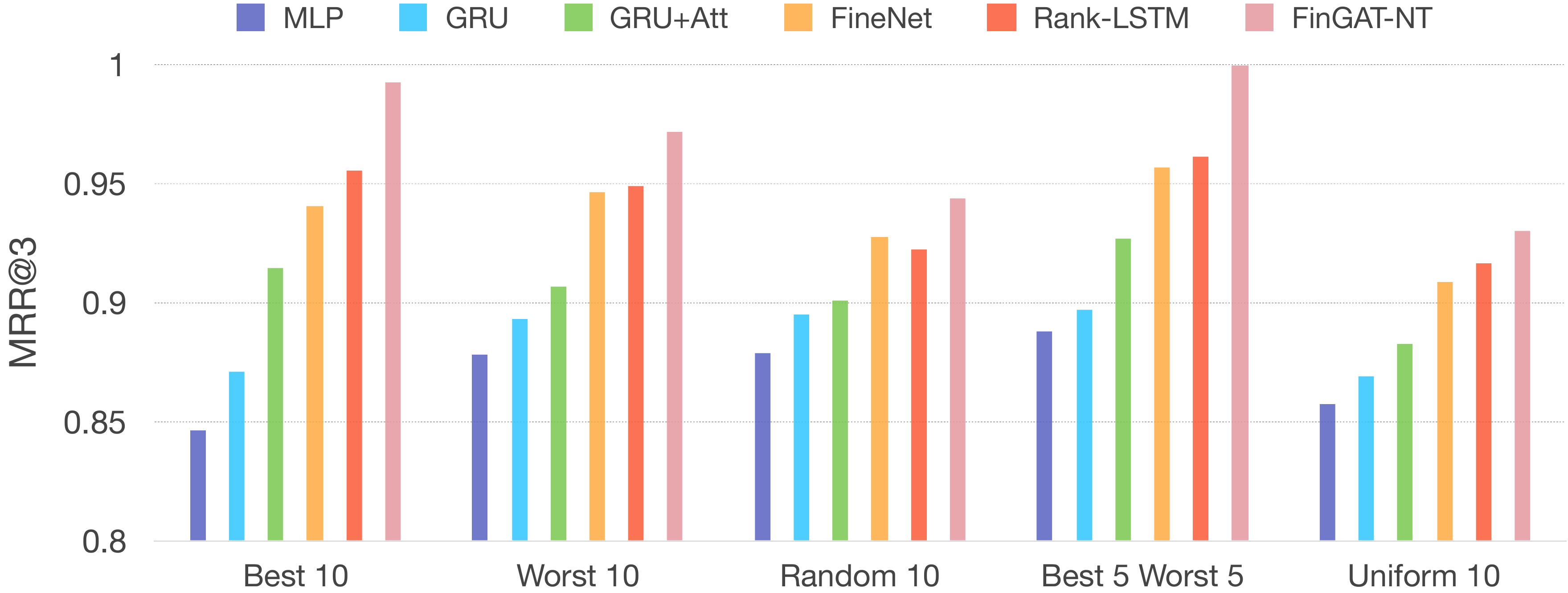}
  \caption{Results on recommendation without sector information in terms of $MRR@3$ for different data subsets of Taiwan Stock dataset.}
  \label{fig:different_data}
\end{figure*}

\textbf{Evaluation without Sector Info.}
We think the effectiveness of \algo~comes from the modeling of stock-sector and sector-sector interactions. However, the sector information is not easily accessible in some stock markets. Besides, we also want to know how \algo~can perform without using sector information. 
Owing to such two reasons, we devise a compact version of \algo~by making some modification, denoted by \algorelation, and conduct an experiment to evaluate its performance on stock recommendation. First, we remove the component of sector-level modeling (Section~\ref{sec:sectorlevel}) as we no longer have sector information. Second, we remove the intra-sector modeling (described in Section~\ref{sec:stocklevel}) that creates a fully-connected graph between stocks belonging to the same sector, i.e., having multiple intra-sector graphs for GAT. Instead, we create a fully-connected graph $G_T$ of stocks, in which all stocks are directly linked. Then we apply the GAT model $G_T$, in which the initial feature vector of each stock node $s_q$ is $\mathbf{a}_i^{s_q}$. After GAT and long-term sequential learning in Section~\ref{sec:stocklevel}, by changing Equation~\ref{finalfv}, we generate the final feature vector $\pi_i^F(s_q)$, given by:
\begin{equation}
\pi_i^F(s_q)=ReLU([\pi_i^G(s_q) \parallel \pi_i^A(s_q)]\mathbf{W}_f).
\end{equation}
Then the same model learning in Section~\ref{sec:modellearn} is applied.

\begin{table*}[!t]
\centering
\caption{Results on ablation study of the proposed \algo.}
\label{tab:ablation}
\resizebox{0.85\textwidth}{!}{%
\begin{tabular}{cccccccc}
\hline
\multicolumn{8}{c}{Taiwan Stock}                                                                                                                                                                                \\ \hline
\multicolumn{1}{c|}{}           & \multicolumn{2}{c|}{$K = 5$}              & \multicolumn{2}{c|}{$K = 10$}             & \multicolumn{2}{c|}{$K = 20$}                                 & Overall                     \\ \cline{2-8} 
\multicolumn{1}{c|}{Model}      & MRR    & \multicolumn{1}{c|}{Precision} & MRR    & \multicolumn{1}{c|}{Precision} & MRR                        & \multicolumn{1}{c|}{Precision} & ACC                         \\ \hline
\multicolumn{1}{c|}{Full model} & $0.4391$ & \multicolumn{1}{c|}{$0.1133$}    & $0.8479$ & \multicolumn{1}{c|}{$0.2022$}    & $1.4106$                     & \multicolumn{1}{c|}{$0.2756$}    & $0.5682$                      \\
\multicolumn{1}{c|}{w/o intra}  & 0.3576 & \multicolumn{1}{c|}{$0.1033$}    & $0.7128$ & \multicolumn{1}{c|}{$0.1317$}    & $1.3406$                     & \multicolumn{1}{c|}{$0.2586$}    & $0.5412$                      \\
\multicolumn{1}{c|}{w/o inter}  & 0.3950 & \multicolumn{1}{c|}{$0.1122$}    & $0.7464$ & \multicolumn{1}{c|}{$0.1511$}    & $1.3887$                     & \multicolumn{1}{c|}{$0.2611$}    & $0.5509$                      \\
\multicolumn{1}{c|}{w/o MTL}    & 0.4215 & \multicolumn{1}{c|}{$0.1127$}    & $0.8023$ & \multicolumn{1}{c|}{$0.1856$}    & $1.4089$                     & \multicolumn{1}{c|}{$0.2723$}    & $0.5342$                      \\
\multicolumn{1}{c|}{w/ MSE}      & $0.3486$ & \multicolumn{1}{c|}{$0.0744$}    & $0.6867$ & \multicolumn{1}{c|}{$0.1228$}    & \multicolumn{1}{l}{$1.1783$} & \multicolumn{1}{c|}{$0.1850$}    & $0.5078$                      \\ \hline
\multicolumn{8}{c}{S\&P 500}                                                                                                                                                                                     \\ \hline
\multicolumn{1}{c|}{}           & \multicolumn{2}{c|}{$K = 5$}              & \multicolumn{2}{c|}{$K = 10$}             & \multicolumn{2}{c|}{$K = 20$}                                 & \multicolumn{1}{l}{Overall} \\ \cline{2-8} 
\multicolumn{1}{c|}{Model}      & MRR    & \multicolumn{1}{c|}{Precision} & MRR    & \multicolumn{1}{c|}{Precision} & MRR                        & \multicolumn{1}{c|}{Precision} & ACC                         \\ \hline
\multicolumn{1}{c|}{Full model} & $0.1974$ & \multicolumn{1}{c|}{$0.0419$}    & $0.3357$ & \multicolumn{1}{c|}{$0.0677$}    & $0.5687$                     & \multicolumn{1}{c|}{$0.0989$}    & $0.5425$                      \\
\multicolumn{1}{c|}{w/o intra}  & $0.1432$ & \multicolumn{1}{c|}{$0.0355$}    & $0.2391$ & \multicolumn{1}{c|}{$0.0500$}    & $0.4282$                     & \multicolumn{1}{c|}{$0.0793$}    & $0.5284$                      \\
\multicolumn{1}{c|}{w/o inter}  & $0.1369$ & \multicolumn{1}{c|}{$0.0301$}    & $0.2382$ & \multicolumn{1}{c|}{$0.0398$}    & $0.4115$                     & \multicolumn{1}{c|}{$0.0877$}    & $0.5371$                      \\
\multicolumn{1}{c|}{w/o MTL}    & $0.1773$ & \multicolumn{1}{c|}{$0.0409$}    & $0.2904$ & \multicolumn{1}{c|}{$0.0581$}    & $0.5033$                     & \multicolumn{1}{c|}{$0.0892$}    & $0.5411$                      \\
\multicolumn{1}{c|}{w/ MSE}      & $0.1072$ & \multicolumn{1}{c|}{$0.0172$}    & $0.2203$ & \multicolumn{1}{c|}{$0.0387$}    & $0.3808$                     & \multicolumn{1}{c|}{$0.0683$}    & $0.5177$                      \\ \hline
\end{tabular}
}
\end{table*}

With \algorelation, it is inevitable to have high computation complexity as the number of stocks increases. 
Therefore, we would suggest the investors to select a subset of stocks to lower down the computational cost of \algorelation~when they request recommendation. Such a setting is realistic because people usually have some candidate stocks in mind when doing investment.
To conduct the experiment, we generate five candidate subsets of stocks by simulating different investment scenarios. 
Specifically, we sort all stocks in Taiwan Stock dataset by their market values~\footnote{\url{https://www.taifex.com.tw/cht/9/futuresQADetail}}, and accordingly generate five stock subsets: 
\begin{itemize}
\item \textbf{``Best 10''}: 10 stocks with the highest market values.
\item \textbf{``Worst 10''}: 10 stocks with the lowest market values.
\item \textbf{``Best 5 Worst 5''}: 5 stocks with the highest market values and 5 stocks with the lowest market values.
\item \textbf{``Random 10''}: randomly selecting 10 stocks from all stocks.
\item \textbf{``Uniform 10''}: dividing the list of sorted stocks into 10 zones, and randomly selecting one stock from each zone. 
\end{itemize}

The results in terms of $MRR@3$ without sector information is shown in
Figure~\ref{fig:different_data}.
It can be found that both the proposed \algorelation~significantly outperforms the competing methods.
Such results indicate that \algorelation~is able to capture the latent interactions between stocks even though no sector information is given. Besides, among the five subsets, the superiority of \textbf{``Best 10''}, \textbf{``Worst 10''}, and \textbf{``Best 5 Worst 5''} is more apparent. Such a result may deliver an interesting insight: the latent relationships between homogeneous stocks in terms of market value are stronger. That said, our \algorelation~can benefit more from capturing the mutual influence between homogeneous stocks.

\textbf{Ablation Study.}
We examine the usefulness of each component in \algo~using Taiwan Stock data. 
We compare the following submodels of \algo, in which the last three remove one component from the complete \algo. 
\begin{enumerate}
\item \textbf{\algo~(Full Model)}: using all components of the proposed \algo.
\item \textbf{w/o intra-sector graph attention (w/o intra)}: \algo~without using embeddings $\tau_i^G(s_q)$ derived from intra-sector graph attention network.
\item \textbf{w/o inter-sector graph attention (w/o inter)}: \algo~without using embeddings $\tau_i(\pi_c)$ obtained from inter-sector graph attention network.
\item \textbf{w/o multi-task leaning (w/o MTL)}: optimizing \algo~solely on pairwise ranking loss.
\item \textbf{with mean square error loss (w/ MSE)}: replacing movement prediction loss (i.e., binary cross entropy loss) with regression loss by mean squared error, where the sigmoid function in Eq.~\ref{eq:task_prediction} is also removed accordingly.
\end{enumerate}

\begin{figure*}[!t]
  \centering
  \includegraphics[width=0.8\linewidth]{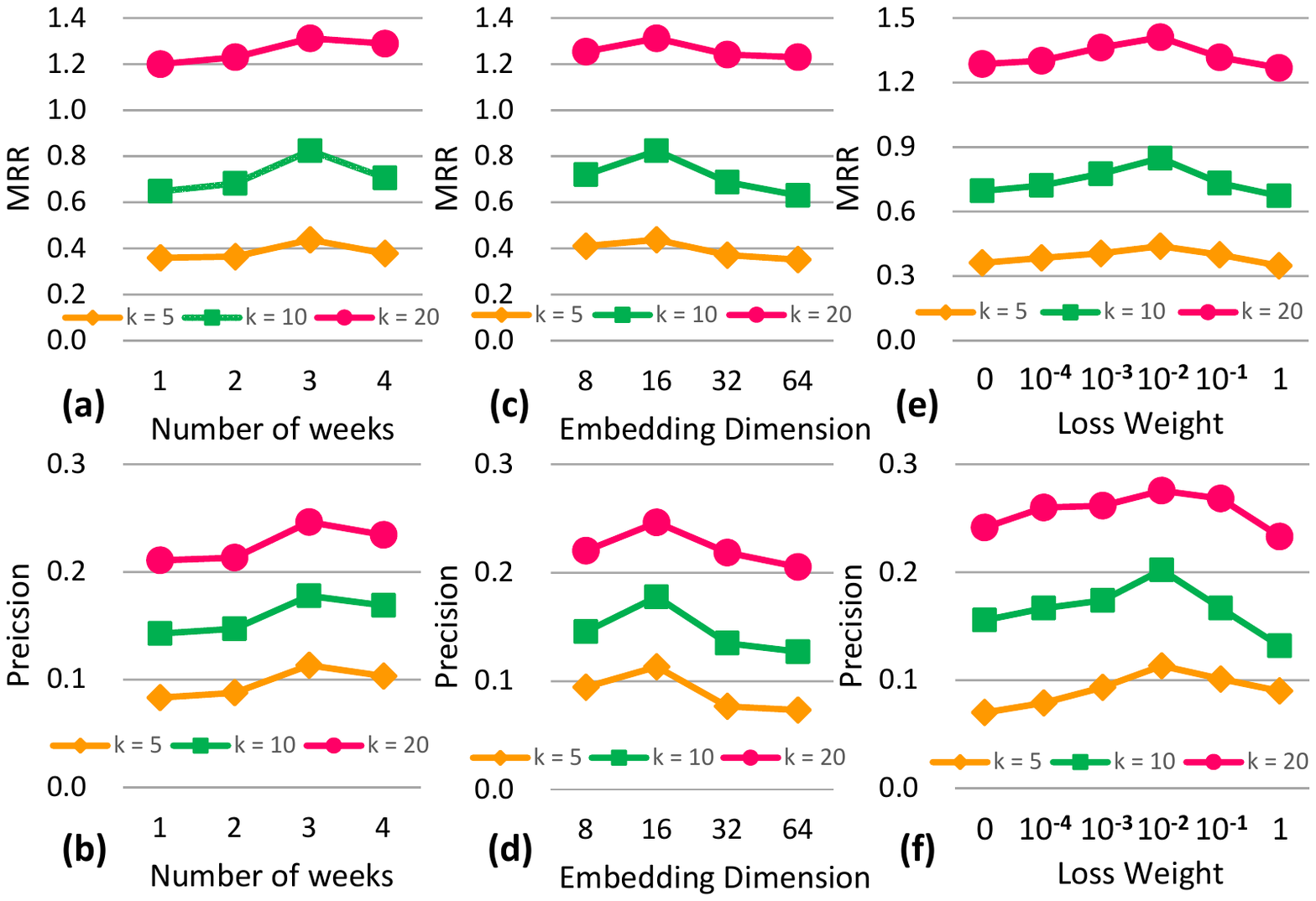}
  \caption{Performance by varing (a)\&(b): the number of training weeks, (c)\&(d): the embedding size, and (e)\&(f): the balance weight $\delta$ in \algo.}
  \label{fig:hyper_analysis}
\end{figure*}

The results are shown in Table~\ref{tab:ablation}. We can find that the full \algo~model leads to the best results in all metrics and settings. Removing any of the three components hurts the recommendation. Such results prove the usefulness of each component. By looking into the three submodels, removing intra-sector modeling leads to the most performance drop, comparing to the other two. This implies that the latent relations between stocks have a direct and significant impact on profitable stock recommendation. In contrast, using only pairwise ranking loss produces the least performance drop. Nevertheless, this result also implies that jointly predicting the stock movement can improve the performance.
To further analyze the effectiveness of binary prediction of stock movement, we implement a variant of \algo~that replaces the movement prediction with a regression prediction task (i.e., \textbf{w/ MSE}). The results show that utilizing MSE loss drastically hurts the performance. We conclude this phenomenon due to the loss curve for squared loss is flatter than binary cross-entropy loss, which leads to training difficulty for optimization.

\subsection{Hyperparameter Analysis}
We aim at understanding how \algo~performs by varying the values of different hyperparameters, including the number of training weeks, the dimension of embeddings used in GRU and GAT, and the balancing parameter $\delta$ that determines the contributions of two tasks. When varying any one of these hyperparameters, we fix other hyperparameters with default settings mentioned in Section~\ref{sec:expsetting}. 


\textbf{Number of training weeks.}
Figure~\ref{fig:hyper_analysis}(a) and \ref{fig:hyper_analysis}(b) shows the performance is affected by changing the input of the number of weeks (i.e., the observed weeks of each stock in the training data). We can find that using at least three weeks of stock time series for training leads to higher MRR and precision scores. few weeks (i.e., $1$ or $2$) for training can result in a bit worse performance. Such results indicate that having both short-term and long-term trends of stocks is essential to better learn their representations for stock recommendation. These results also correspond to that incorporating both short-term and long-term information can produce better performance by the two stronger competing methods FineNet~\cite{tsai2019finenet} and RankLSTM~\cite{feng2019temporal}, as shown in Table~\ref{tab:main_result}. Nevertheless, the proposed \algo~is better than FineNet and RankLSTM because we not only learn stock embeddings from short- and long-term information, but also exploit them to model the latent interactions between stocks and between sectors. 

\textbf{Embedding Size.}
The results exhibited in Figure~\ref{fig:hyper_analysis}(c) and \ref{fig:hyper_analysis}(d) shows that when the embedding size $=16$ leads to better performance. Too small ($8$) or too large values ($32$ or $64$) of embedding size worsen the recommendation quality. The possible reason could be underfitting and overfitting, respectively. Hence, we would suggest to use the embedding size $16$ in \algo.


\textbf{Balancing Parameter $\delta$.}
We present the performance by varying $\delta \in \{0,0.0001,0.001,0.01,0.1,1\}$ in Figure~\ref{fig:hyper_analysis}(e) and \ref{fig:hyper_analysis}(f).
It can be found that $\delta=0.01$ leads to the best performance. A small $\delta$ indicates less contribution of cross-entropy loss in stock movement prediction and much contribution of pairwise ranking loss in profitable stock recommendation. 
We can also find that when we consider only either the loss of stock movement prediction (i.e., $\delta = 1.0$) or the loss of profitable stock recommendation (i.e., $\delta = 0.0$), the performance gets much worsened. The results imply that properly fusing two tasks can improve the performance. Based on the results, we suggest to set $\delta=0.01$ for \algo.


\begin{figure}[!t]
\centering  
\subfigure[``Consumer \& Goods'' Sector]{
\label{fig:intra_consumer_sector}
\includegraphics[width=0.85\linewidth]{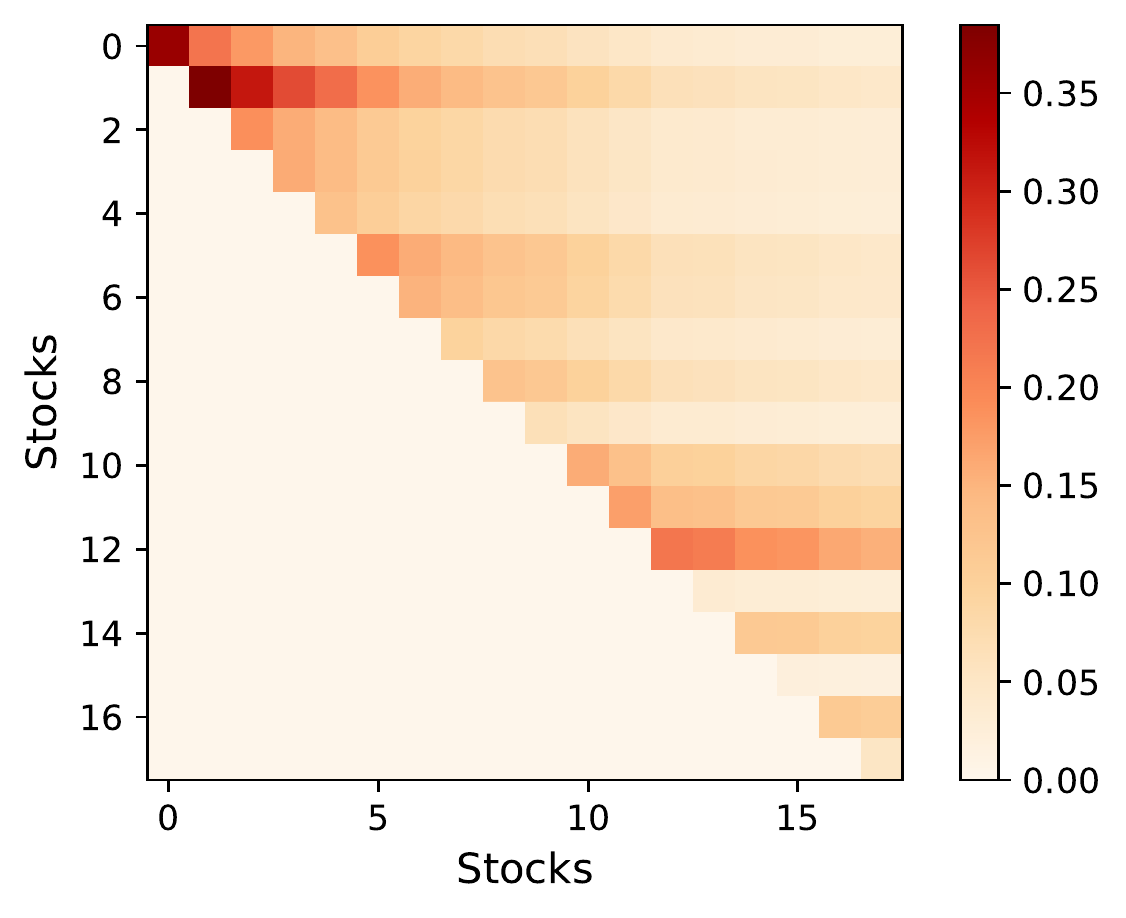}}
\subfigure[``Energy'' Sector]{
\label{fig:intra_energy_sector}
\includegraphics[width=0.85\linewidth]{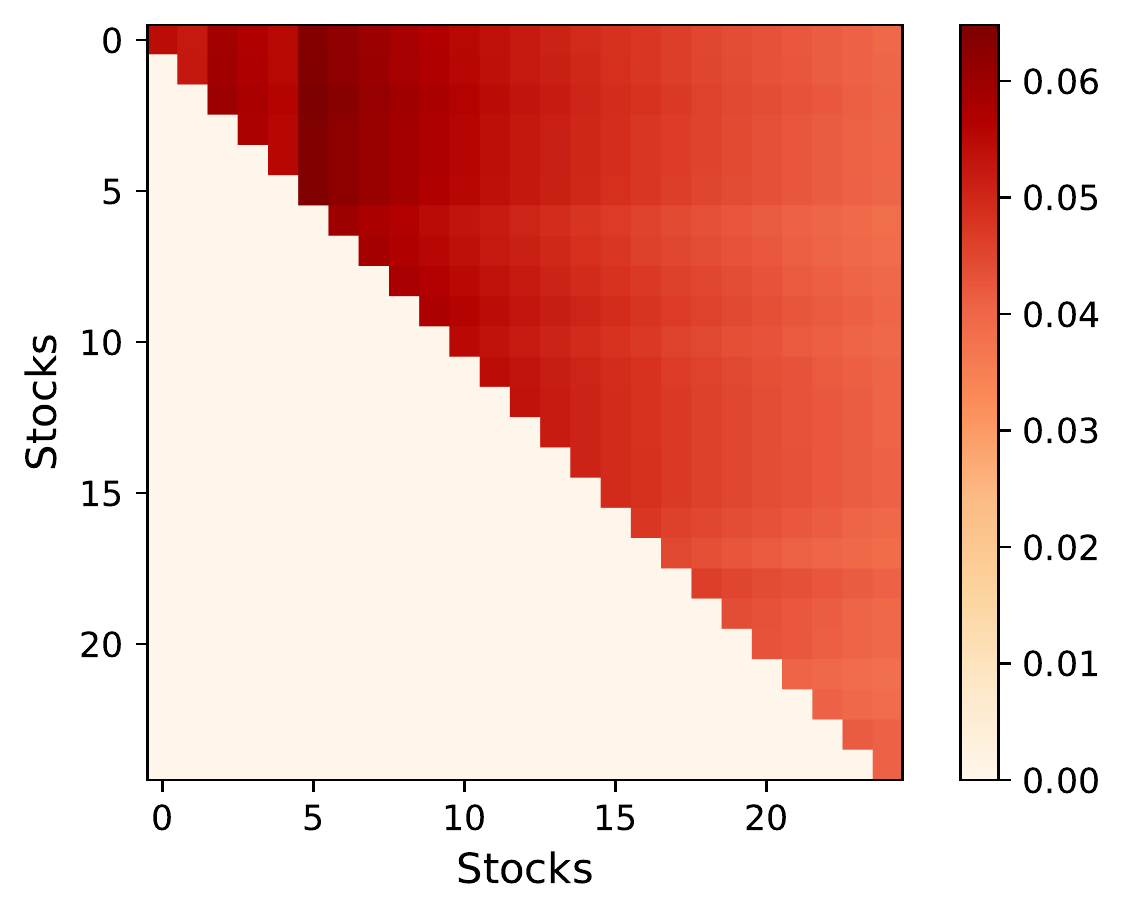}}

\caption{Visualization of attention weights between same-sector stocks.}
\label{fig:attention_weights1}
\end{figure}


\begin{figure}[!t]
\centering  
\subfigure[Inter-sector Relationship on Taiwan Stock]{
\label{fig:inter_sector_tw}
\includegraphics[width=1.0\linewidth]{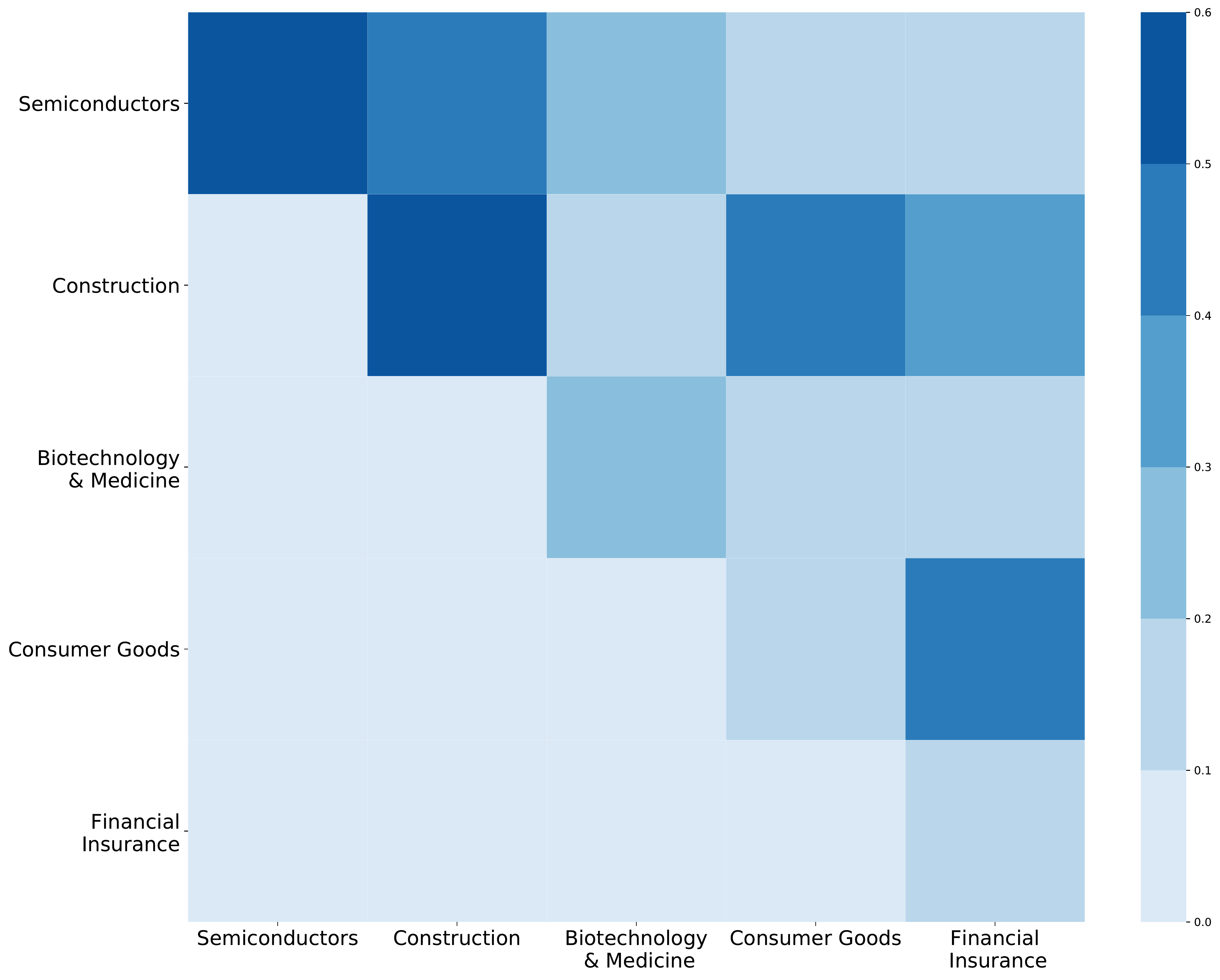}}
\subfigure[Inter-sector Relationship on S\&P 500]{
\label{fig:inter_sector_sp}
\includegraphics[width=1.0\linewidth]{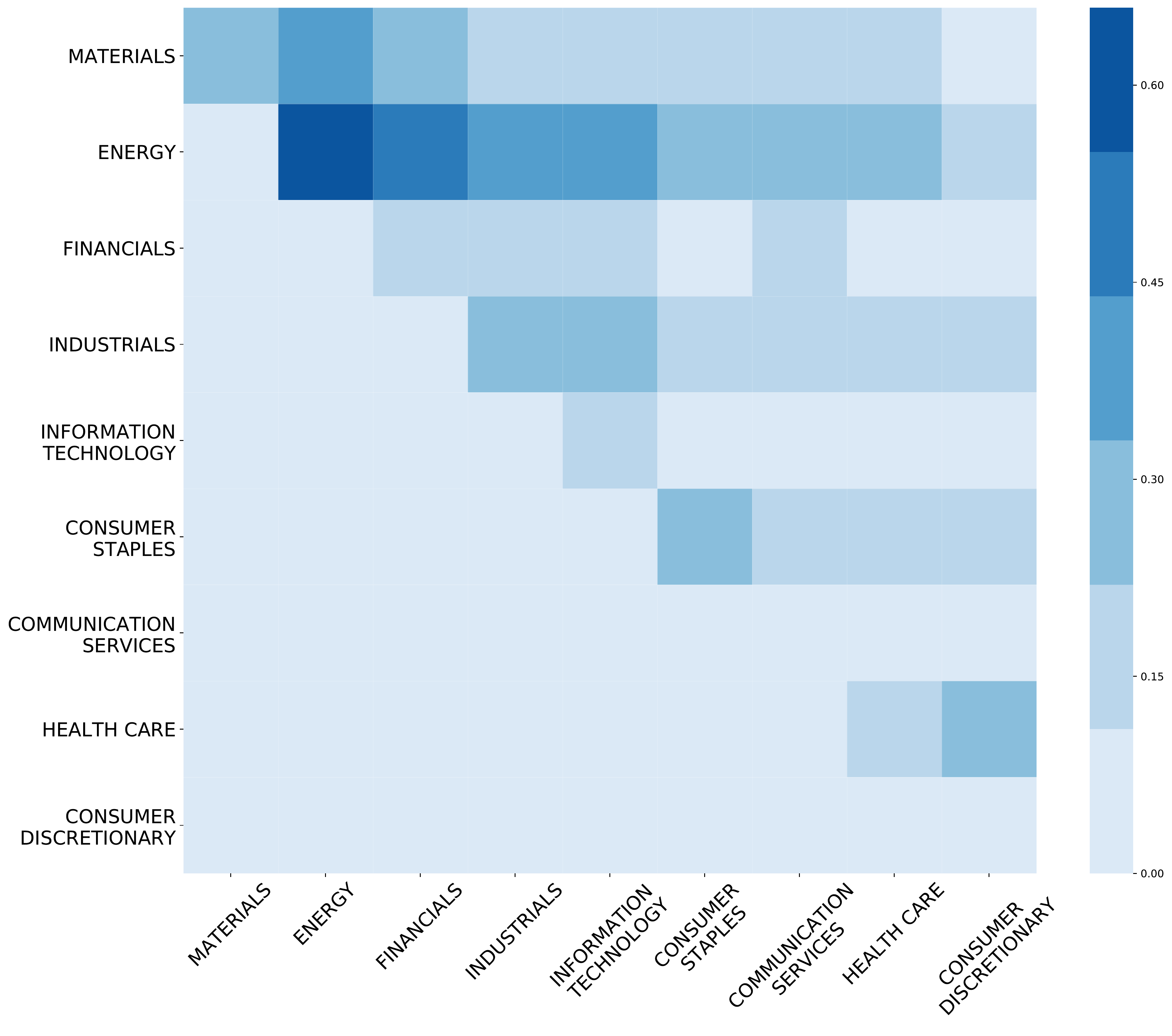}}

\caption{Visualization of inter-sector attention weights.}
\label{fig:attention_weights2}
\end{figure}




\subsection{Exploring Latent Internations via Attention Weights}
\algo~aims to learn latent interactions in two aspects: intra-sector and inter-sector. As described in Section~\ref{sec:stocklevel}, the intra-sector graph attention network models the interactions between same-sector stocks, e.g., 
business competition relations or upstream/downstream companies in the real-world stock market. 
The inter-sector graph attention network (Section~\ref{sec:sectorlevel}) further learns the sector-sector correlation to capture the global trend influence between different economic industries, e.g., 
the influence bteween oil, gold, and textile sectors,
whish are discussed in Section~\ref{sec-intro} and Figure~\ref{fig:Toy_example}.
It is worthwhile and essential to explore the latent intra-sector and inter-sector relations learned by attention weights since
\algo~exploits relation learning via graph attention networks. 

\textbf{Intra-sector Attention.} 
We present the attention weights (on intra-sector modeling) of a randomly-selected testing instance, i.e., same-sector stock-stock attention weights. 
The attention visualization plots for the ``Consumer \& Goods'' sector in Taiwan stock data and the ``Energy'' sector in S\&P 500 data are exhibited
as exhibited in Figure~\ref{fig:intra_consumer_sector} and~\ref{fig:intra_energy_sector}, respectively.
By looking into Figure~\ref{fig:intra_consumer_sector}, we can find that the cells of high stock-stock attention weights are distributed into subgroups (i.e., stock IDs $0$ to $3$ and stock IDs $10$ to $12$). Such results indicate that stock prices in the ``Consumer \& Goods'' sector are highly correlated with or mutually influenced to each other in these subgroups. Since the ``Consumer \& Goods'' sector contains stock categories manufacturers, retailers, and distributors, and they often intensely collaborate to form some supply chains~\cite{stadtler2002supply}, we think it is reasonable to have subgroups of stocks whose prices are correlated. 
As for the visualization plot of the ``Energy'' sector in Figure~\ref{fig:intra_energy_sector}, it is apparent that
the attention weights between stocks tend to be uniformly distributed.
These results imply that the interaction effect between stocks in the ``Energy'' sector is relatively significant, comparing to Figure~\ref{fig:intra_consumer_sector}, even though the attention values are low. This phenomenon may be due to the scarcity of energy resources that leads to an intense competition (e.g., oil price war~\footnote{\url{https://en.wikipedia.org/wiki/2020_Russia\%E2\%80\%93Saudi_Arabia_oil_price_war}}) and high fluctuation. A study~\cite{broadstock2016shocks} also showed the impact of oil shock could substantially lead to gasoline price shocks. 
In short, in addition to profitable stock recommendation, we believe that the sector-sector attention weights generated by our \algo~model can provide insights on stock influence and correlation to help the decision making of investors.

\textbf{Inter-sector Attention.}
We present the attention weights of inter-sector modeling, i.e., sector-sector attention weights, of a randomly-selected testing instance. The plots generated from Taiwan stock and S\&P 500 datasets are demonstrated in Figure~\ref{fig:inter_sector_tw} and Figure~\ref{fig:inter_sector_sp}, respectively.
In Figure~\ref{fig:inter_sector_tw}, aside from the diagonal, we can find that the attention weight between ``Semiconductors'' and ``Construction'' sectors is relatively higher than other sector-sector cells. Besides, both ``Semiconductors'' and ``Construction'' sectors have a relatively higher impact on other sectors. The possible reason lies in that both are the fundamentals of Taiwan's Economy~\footnote{\url{https://en.wikipedia.org/wiki/Economy_of_Taiwan}}.
Sectors between ``Consumer Goods'' and ``Financial Insurance'' also contribute significantly since they all regard supply and demand through various financial behaviors.
On the other hand, in Figure~\ref{fig:inter_sector_sp} for S\&P 500 data, the ``Energy'' sector has higher a correlation with different sectors. This is reasonable since ``Energy'' sector involves oil, gasoline, and fossil fuel industries that have been shown to bring a critical impact on the stock market~\cite{sine2009tilting,phan2015stock}. Note that \algo~learns the strong influence of the ``Energy'' sector without any prior knowledge. This verifies our discussion in Figure~\ref{fig:Toy_example} that the latent relations between sectors can be learned via graph neural networks. 

\begin{figure}[!t]
\centering  
\subfigure[Distributions of attention weights]{
\label{fig:attetion_score}
\includegraphics[width=0.85\linewidth]{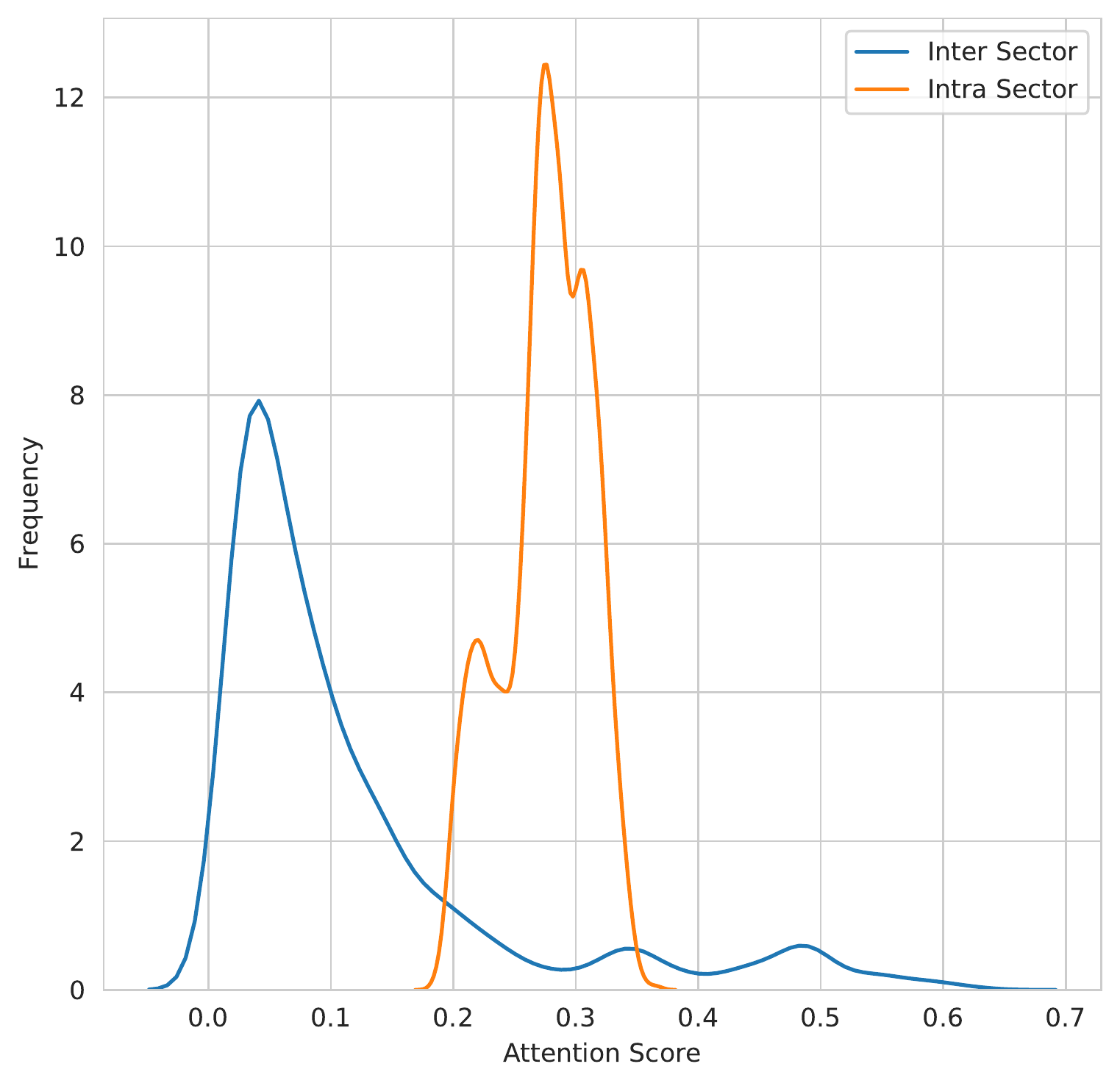}}
\subfigure[Variance distributions of attention]{
\label{fig:attetion_var} 
\includegraphics[width=0.85\linewidth]{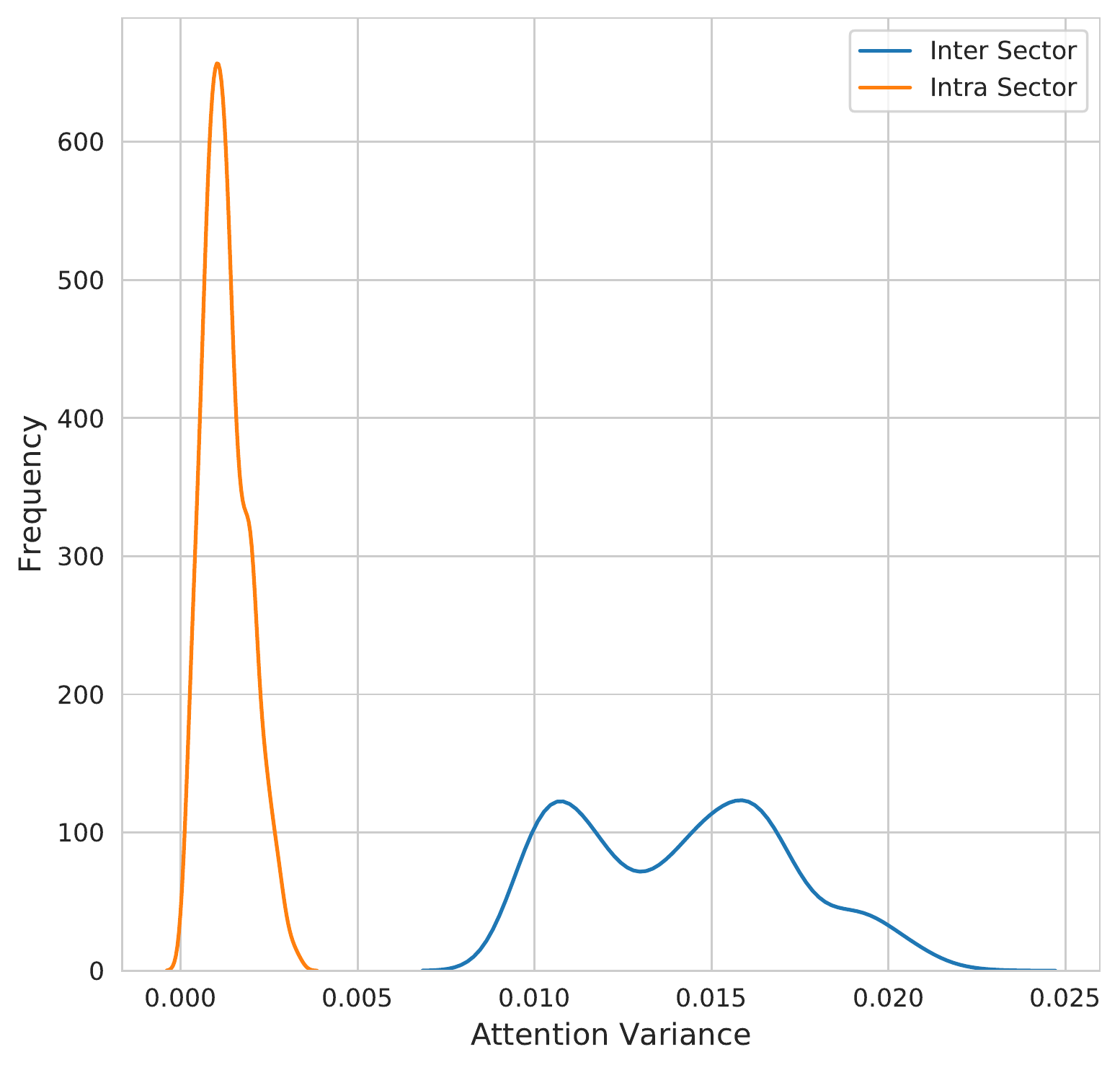}}
\caption{Distributions and varances of attention weights}
\label{fig:attention_distribution}
\end{figure}

\textbf{Distributions of Attention Weights.}
To further explore the differences between intra-sector and inter-sector, we collect attention weights from intra-sector and inter-sector GATs over all testing instances, and demonstrate their distributions in Figure~\ref{fig:attention_distribution} using Taiwan Stock dataset. 
We first present the distribution of attention weights in Figure~\ref{fig:attetion_score}. The inter-sector attention weights (blue) tend to have a right-skewed distribution. This means that only a few sectors are highly correlated with each other. Such a result reveals that the stock market could be possibly dominated by the minority leading sectors (e.g., the ``Semiconductor'' sector in Figure~\ref{fig:inter_sector_tw}). The stock-stock attention weights (orange) are around located in $(0.2,0.4)$ interval.
This indicates that there are no significantly influential stocks within any sector that contribute large impact to other stocks. Figure~\ref{fig:attetion_var} further shows the variance of attention weights (computed over testing instances). Inter-sector attention weights have a higher variance than intra-sector ones. The results again confirm that leading sectors would drastically influence other sectors while stocks within a sector do not exhibit strong dependency.

\section{Conclusion}
\label{sec-conclude}
This work aims at recommending the most profitable stocks using price time series and sector information of stocks. We develop a novel deep learning-based model, \algo, to achieve the goal. The novelty of FinGAT is three-fold. First, FinGAT requires no pre-defined relationships between stocks, comparing to existing studies that presume the relationships between stocks are given. We exploit graph attention networks to automatically learn the latent interactions between stocks and sectors. Second, the two-level hierarchy that depicts the belonging between stocks and sectors is used to pass information so that both fine-grained and coarse-grained latent relationships can be modeled. Third, FinGAT leverages a multi-task objective that jointly optimizes the loss functions of profitable stock recommendation and stock movement prediction. We conduct experiments using Taiwan Stock, S\&P 500, and NASDAQ datasets. The results show that \algo~can significantly outperform the state-of-the-art and baseline competing methods. \algo~is also able to generate promising recommendation performance without using sector information.

The future extension based on our \algo~framework is three-fold. First, while we currently construct fully-connected graphs to depict the interactions between stocks and between sectors, we argue that there exists a better structure that captures how stocks/sectors are influenced by each other. Hence, we will try to incorporate the inference of graph structures into \algo~as a joint learning mechanism. Second, listed companies usually contain metadata that provides fine-grained attributed descriptions. We aim at creating a knowledge graph based on such metadata so that the correlation between stocks can be better encoded into embeddings. Third, since stock prices are sensitive to daily news, we aim at learning the representation from stock-related news, and utilize accordingly to further boost the \algo~recommendation performance.

\ifCLASSOPTIONcompsoc
  \section*{Acknowledgments}
\else
  \section*{Acknowledgment}
\fi

This work is supported by Ministry of Science and Technology (MOST) of Taiwan under grants 109-2636-E-006-017 (MOST Young Scholar Fellowship) and 109-2221-E-006-173, and also by Academia Sinica under grant AS-TP-107-M05.

\ifCLASSOPTIONcaptionsoff
  \newpage
\fi



%
\bibliographystyle{plain}
\bibliography{reference}

\begin{IEEEbiography}[{\includegraphics[width=1in,height=1.25in,clip,keepaspectratio]{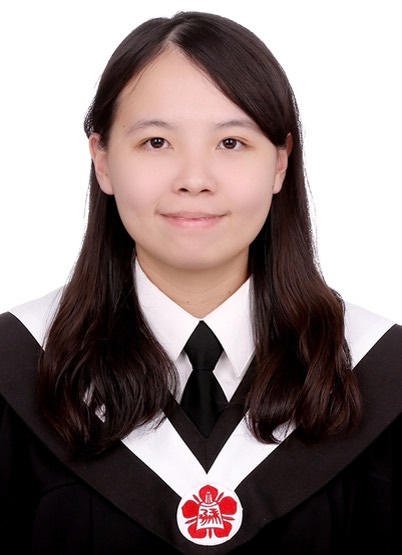}}]%
{Yi-Ling Hsu} is now a graduate student in Master Program in Statistics, National Taiwan University (NTU), Taipei, Taiwan. She received a Bachelor's degree in Statistical Science from National Cheng Kung University (NCKU), Tainan, Taiwan, in 2020. She is also a research assistant in Networked Artificial Intelligence Laboratory at NCKU. Her main research interests include Data Mining, Machine Learning, Financial Technology, and Statistics.
\end{IEEEbiography}
\begin{IEEEbiography}[{\includegraphics[width=1in,height=1.25in,clip,keepaspectratio]{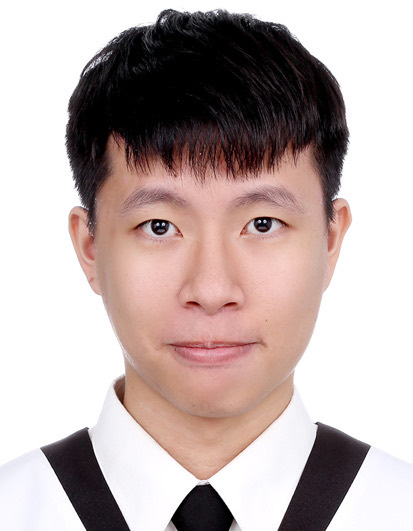}}]%
{Yu-Che Tsai} is now a Master graduate student in Department of Computer Science and Information Engineering, National Taiwan University (NTU), Taipei, Taiwan. He received a Bachelor's degree in Statistical Science from National Cheng Kung University (NCKU), Tainan, Taiwan, in 2020. He is also a research assistant in Networked Artificial Intelligence Laboratory at NCKU. His main research interests include Data Mining, Machine Learning, Recommender Systems, and Deep Learning. His papers had been published in KDD 2020, CIKM 2019, and RecSys 2019.
\end{IEEEbiography}
\begin{IEEEbiography}[{\includegraphics[width=1in,height=1.25in,clip,keepaspectratio]{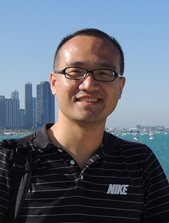}}]%
{Cheng-Te Li} is an Associate Professor at Institute of Data Science and Department of Statistics, National Cheng Kung University (NCKU), Tainan, Taiwan. He received my Ph.D. degree (2013) from Graduate Institute of Networking and Multimedia, National Taiwan University. Before joining NCKU, He was an Assistant Research Fellow (2014-2016) at CITI, Academia Sinica. Dr. Li's research targets at Machine Learning, Deep Learning, Data Mining, Social Networks and Social Media Analysis, Recommender Systems, and Natural Language Processing. He has a number of papers published at top conferences, including KDD, WWW, ICDM, CIKM, SIGIR, IJCAI, ACL, EMNLP, NAACL, RecSys, and ACM-MM. He leads Networked Artificial Intelligence Laboratory (NetAI Lab) at NCKU.
\end{IEEEbiography}



%








\end{document}